\documentclass{article} % For LaTeX2e
\usepackage{iclr2027_conference,times}

\usepackage{amsmath,amsfonts,bm}

\def\eqref#1{equation~\ref{#1}}
\def\1{\bm{1}}

\DeclareMathAlphabet{\mathsfit}{\encodingdefault}{\sfdefault}{m}{sl}
\SetMathAlphabet{\mathsfit}{bold}{\encodingdefault}{\sfdefault}{bx}{n}

\usepackage{hyperref}
\usepackage{url}
\usepackage{booktabs}       % professional-quality tables
\usepackage{amsfonts}       % blackboard math symbols
\usepackage{nicefrac}       % compact symbols for 1/2, etc.
\usepackage{microtype}      % microtypography
\usepackage{multirow}
\usepackage{graphicx}
\usepackage{wrapfig}
\usepackage{amsmath}
\usepackage{subcaption}
\usepackage[table]{xcolor}
\usepackage{colortbl}

\definecolor{gaincolor}{RGB}{0,128,0}
\definecolor{losscolor}{RGB}{200,0,0}
\definecolor{rowshade}{gray}{0.93}

\newcommand{\up}[1]{{\tiny\textcolor{gaincolor}{$\uparrow$#1}}}
\newcommand{\dn}[1]{{\tiny\textcolor{losscolor}{$\downarrow$#1}}}

\renewcommand{\headrulewidth}{0pt}

\title{Learning Reliable GUI Agents under Imperfect Priors}

\author{Bo Han$^{1}$, Qianyi Wang$^{1}$, Shuai Liu$^{1}$, Xiong Zifan$^{1}$ \\
\textbf{Changqiao Wu$^{2}$, Yuanfa Li$^{2}$, Pengzhi Gao$^{2}$, Wei Liu$^{2}$, Jian Luan$^{2}$, Heng Qu$^{2}$}\\
\textbf{Yunpeng Song$^{1}$, Zhongmin Cai$^{1}$} \\
$^{1}$Xi'an Jiaotong University \\
$^{2}$MiLM Plus, Xiaomi Inc.
}

\iclrfinalcopy % Uncomment for camera-ready version, but NOT for submission.
\begin{document}

\maketitle

\begin{abstract}
GUI agents built on large language and vision-language models still struggle on unseen applications and complex multi-step tasks, as completing real GUI tasks depends on app-specific, temporally volatile operational knowledge that is scarce in pretraining corpora. Retrieval-augmented execution offers a natural remedy but faces two coupled bottlenecks: knowledge at scale is hard to acquire, and self-collected priors inevitably drift from the live environment due to version updates, promotions, ads, A/B tests, and personalization. We therefore argue that GUI agents should not pursue perfect knowledge but learn to act correctly under imperfect priors, and propose our framework that couples knowledge acquisition with noise-robust utilization: a structured exploration strategy traverses interactive elements, builds a UI state-transition graph, and synthesizes (task, trajectory) pairs via a VLM without human annotation; a noise-aware training strategy, grounded in a taxonomy of real GUI drift patterns, injects five types of realistic errors into self-explored trajectories to teach the agent to assess prior reliability before acting. Experiments on physical devices and online emulator benchmarks show that our method discovers more unique screens, covers more benchmark tasks, and more effectively rejects erroneous priors while leveraging correct ones, with accuracy gains that transfer across datasets.
\end{abstract}

\section{Introduction}

GUI agents built on large language models and vision-language models have made rapid progress on mobile and desktop automation~\citep{cogagent,seeclick,mobile_agent,appagent,ferret_ui,uitars,osatlas}, but they remain brittle on unseen applications and long-horizon tasks~\citep{android_world,android_lab,mobile_world,osworld}. A single screenshot is rarely enough. Completing a real task often depends on app-specific, temporally volatile operational knowledge, such as hidden entry points, preconditions that make a setting visible, or whether a salient button is actually an advertisement. Such knowledge is scarce in pretraining corpora and hard to infer from pixels alone. Retrieval-augmented execution~\citep{rag} offers a natural remedy by retrieving historical trajectories at decision time~\citep{mapagent,moba,mobilegpt,mobile_agent_rag}, yet instantiating it on GUIs raises two coupled and still-open problems: how to \emph{acquire} such knowledge at scale, and how to \emph{use} it when it is imperfect.\looseness=-1

\paragraph{Bottleneck 1: scalable acquisition.} High-quality trajectories are typically collected through manual annotation or user demonstrations~\citep{android_in_the_wild,mind2web,android_lab}, both of which are expensive and unable to cover the long tail of apps. Autonomous exploration provides an alternative~\citep{autodroid,droidagent,droidbot,llm_explorer}, but it comes with structural limitations. Without a user intent to anchor behavior, control-semantics annotations are partly inferential~\citep{screen2words,widget_captioning}; exploration operates under a finite budget; and login states, server-side routing, and personalized content are largely beyond reach. Automatically acquired GUI knowledge should therefore be treated not as ground truth, but as \emph{self-explored priors}: potentially incomplete, outdated, or noisy hints whose applicability agents must assess against the current task and execution state.

\begin{wrapfigure}{R}{0.34\textwidth}
\centering
\includegraphics[width=0.34\textwidth]{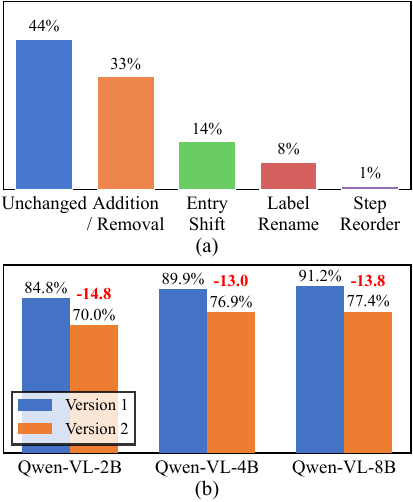}
\caption{(a) GUI drift patterns across app versions; (b) task success drops on newer versions.}
\label{fig:drift}
\end{wrapfigure}

\paragraph{Bottleneck 2: reliability under GUI drift.} Even high-quality priors become unreliable over time, since GUIs drift across multiple scales.  To characterize this phenomenon, we curate 176 tasks from 20 popular apps that span paired old and new versions, and manually compare the completion trajectories across versions. As summarized in Figure~\ref{fig:drift}(a), only 44\% of the tasks retain identical trajectories after an update, while the remainder fall into a small, recurring set of drift patterns: step addition or removal, entry-point or path shift, control renaming, and goal-mismatched or distracting detours. These patterns are consistent with the 74.6\% of paired screens across versions that show visual differences~\citep{lu2025transbench}. Beyond versioning, promotional campaigns, ads, A/B testing, and regional routing continue to reshuffle interfaces. Figure~\ref{fig:drift}(b) quantifies the cost of ignoring this property: even when supplied with correct trajectories from older versions, three competitive fine-tuned models still drop noticeably on newer versions, echoing the well-documented instability of naively appending retrievals in RAG~\citep{robust_irrelevant_context,raat,instructrag,power_of_noise}.

We cast GUI automation as sequential decision making under retrieval-augmented, \emph{imperfect} priors. The agent is conditioned on a path-structured prior $\pi$ that may deviate from the unobserved oracle $\pi^\star$ along any of the drift axes above. Rather than chasing perfect knowledge, we train the agent to \emph{act correctly under imperfect priors}. We propose \textbf{SAGE} (\emph{Self-explored, Assessment-Guided Execution}), which combines three components to support this goal. \emph{(i) Self-explored priors.} Building on UI-graph construction~\citep{autodroid,droidbot}, the agent traverses interactive elements on real devices with semantic-region budgeting and recovery-aware backtracking, and uses a multimodal annotator to synthesize tasks from reachable paths. This procedure yields retrievable $(g,\pi^\star)$ pairs at scale without human involvement. \emph{(ii) Drift-aware perturbation.} Guided by the taxonomy in Figure~\ref{fig:drift}(a), a family of operators $\{T_k\}$ maps clean priors to a mixture $\mathcal{D}_\Pi$ that approximates deployment-time retrieval noise, with each operator paired with a deterministic reliability label. \emph{(iii) Reliability-gated policy.} The policy factorizes as $p_\theta(a\!\mid\!\cdot)=\sum_{r}p_\theta(r\!\mid\!\cdot)\,p_\theta(a\!\mid\!\cdot,r)$ with $r\!\in\!\{\textsc{Follow},\textsc{Partial},\textsc{Ignore}\}$, so the agent commits to an explicit reliability decision before acting. In this way, it learns to follow reliable priors, override conflicting ones, and selectively use partial ones~\citep{selfrag,crag}.

\paragraph{Contributions.} (1) We present an annotation-free exploration pipeline that constructs a deduplicated UI transition graph and yields task-aligned retrievable priors at scale. (2) We introduce a reliability-gated policy trained on a drift-aware prior distribution $\mathcal{D}_\Pi$, which turns retrieval noise into explicit supervision and forces an explicit \textsc{Follow}/\textsc{Partial}/\textsc{Ignore} decision before acting. (3) We construct a benchmark of clean and systematically perturbed $(\text{task},\text{trajectory})$ pairs for retrieval-augmented GUI agents. Upon acceptance, we will publicly release the core implementation and the public-data portion of the benchmark.

% , including processed CMGUI and ChiM-Nav JSON splits, image manifests, prompt templates, representative training configurations, preprocessing scripts, and evaluation scripts. We will not release privacy- or license-sensitive self-explored CUTG artifacts, including raw screenshots, complete trajectories, account states, raw device logs, credentials, server addresses, or local machine metadata.

\section{Related Work}

\textbf{GUI agents and online benchmarks.} Multimodal foundation models have moved GUI agents from static interface understanding to real software interaction, from single-step grounding~\citep{seeclick,cogagent} to mobile agents tuned or reinforced for multi-step execution~\citep{agentcpm_gui,mai_ui,gui_owl,step_gui,mobizen_gui}, including the Qwen3-VL backbones we adopt~\citep{qwen3vl}. Realistic evaluation is enabled by reproducible Android environments with state-based checking~\citep{android_world,android_lab,mobile_world} and by their Chinese-app counterparts~\citep{cmgui,chim_nav}. These works primarily ask whether an agent can complete a task in a given environment; we ask how app-specific operational knowledge can be acquired at scale and used robustly when it is unreliable.\looseness=-1

\textbf{App exploration as a knowledge source.} Autonomous exploration has long served mobile testing and GUI data collection, from UI-guided event generation~\citep{droidbot} to LLM-guided semantic exploration that reduces invalid interactions~\citep{droidagent,llm_explorer}. Raw traces, however, are highly redundant: transient UI changes create spurious states, while functionally equivalent controls are repeatedly explored, yielding fragmented trajectories rather than a connected graph with shared states. We therefore consolidate traces through state deduplication, control-function merging, and transition alignment, treating the resulting transitions and paths as \emph{priors} that can be audited, perturbed, and used for robust learning.

\textbf{Retrieval augmentation for GUI agents.} Retrieval-augmented generation is standard for knowledge-intensive tasks~\citep{rag}. In GUI agents, analogous ideas appear as app memory, trajectory memory, and page- or task-level knowledge bases: explore–select–derive–recall subtask memory~\citep{mobilegpt}, automated dynamic analysis for app-specific knowledge~\citep{autodroid}, and exploration- or demonstration-based operation knowledge~\citep{appagent}. For long-horizon planning, trajectory and multifaceted experience memories~\citep{mapagent,moba} and dual-path Manager–Operator retrieval~\citep{mobile_agent_rag} demonstrate the value of external priors. Prior work mainly focuses on knowledge acquisition and retrieval, often assuming retrieved content is usable. Our work instead targets unreliable priors and interactive noise arising from changing interface states and action consequences.

\section{Method}

\begin{figure}[tbp]
    \centering
    \includegraphics[width=0.8\linewidth]{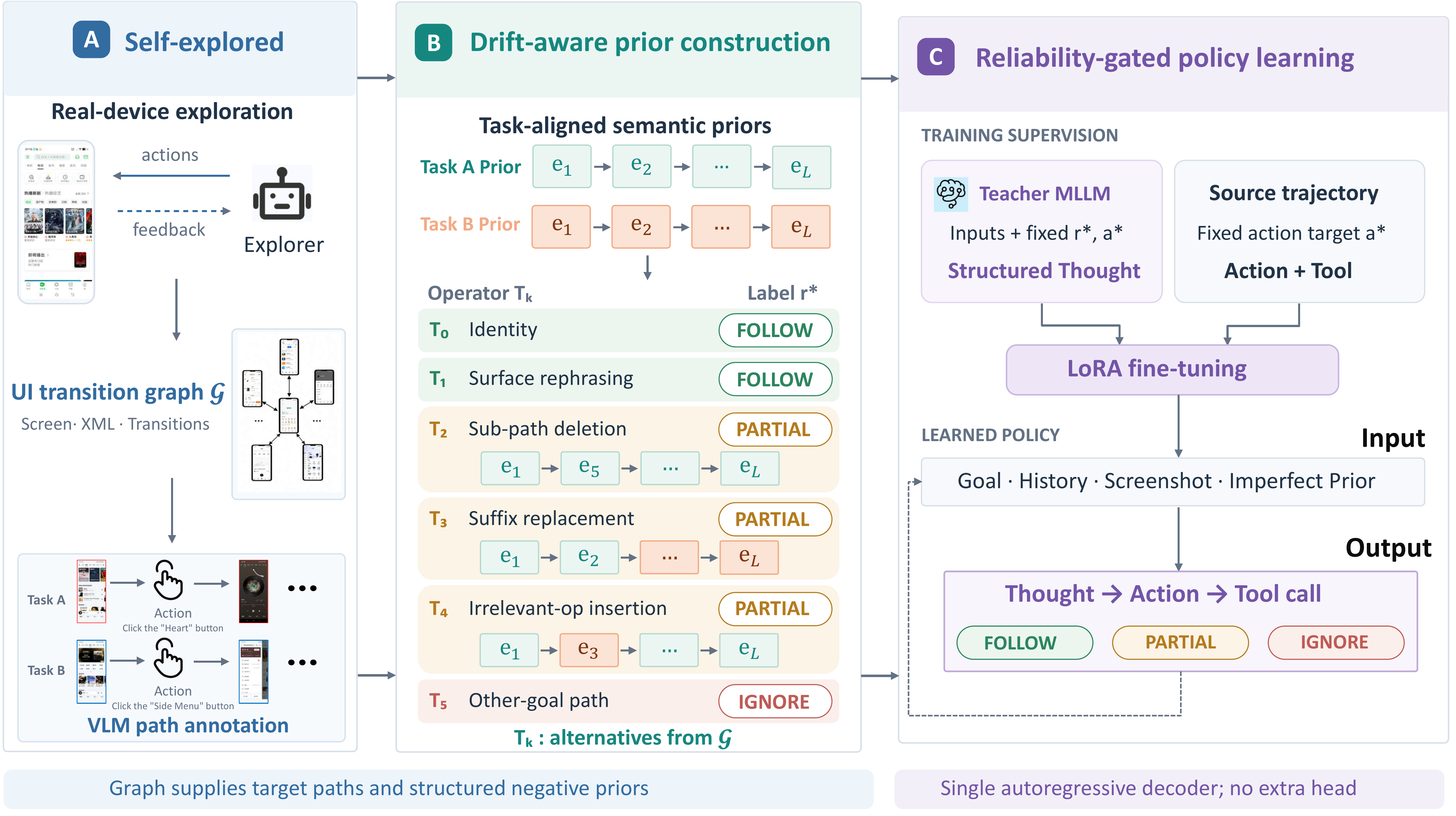}
    \caption{Overview of \textsc{SAGE}. Semantic-guided exploration builds a deduplicated UI graph $\mathcal{G}$ and task-aligned priors. Drift-aware operators $\{T_k\}$ construct $\mathcal{D}_\Pi$ with reliability labels. Teacher-generated thoughts supervise a LoRA-fine-tuned policy that assesses priors as \textsc{Follow}, \textsc{Partial}, or \textsc{Ignore} before acting.}
    \label{fig:framework}
\end{figure}

% We cast GUI automation as sequential decision making under retrieval-augmented, \emph{imperfect} priors. \textsc{SAGE} comprises three coupled components: (i) a human-free mechanism that produces operational knowledge (Sec.~\ref{sec:exploration}); (ii) a drift-aware shaping of the training prior distribution grounded in an empirical taxonomy of GUI drift (Sec.~\ref{sec:perturb}); and (iii) a reliability-gated policy that commits to an explicit reliability decision before acting (Secs.~\ref{sec:formulation}--\ref{sec:objective}).
\textsc{SAGE} combines self-explored prior acquisition (Sec.~\ref{sec:exploration}), drift-aware perturbation (Sec.~\ref{sec:perturb}), and reliability-gated action generation (Secs.~\ref{sec:formulation} and~\ref{sec:objective}), as shown in Figure~\ref{fig:framework}.

\subsection{Problem Formulation}
\label{sec:formulation}

\paragraph{GUI decision process.} We model a GUI task as a contextual decision process $(\mathcal{S},\mathcal{A},\mathcal{G},\mathcal{T},\rho)$, with observable states $\mathcal{S}$ (screenshot plus accessibility metadata), executable actions $\mathcal{A}$, natural-language goals $\mathcal{G}$, a non-stationary transition kernel $\mathcal{T}:\mathcal{S}\!\times\!\mathcal{A}\!\to\!\Delta(\mathcal{S})$ induced by the live app, and a completion indicator $\rho$. At step $t$, the agent observes $s_t$, conditions on goal $g$ and history $h_{<t}=(s_{<t},a_{<t})$, and emits $a_t\!\in\!\mathcal{A}$.

\paragraph{Retrieval-augmented, imperfect priors.} The agent is given a path-structured prior $\pi=(\bar e_1,\dots,\bar e_L)\!\in\!\Pi$ retrieved from an external store, where each $\bar e_\ell$ narrates a pre-state, triggering control, and post-state in natural language, yielding a policy $p_\theta(a_t\mid s_t,h_{<t},g,\pi)$. Let $\pi^{\star}(g)$ be the (unobserved) oracle path for $g$ in the live environment. A real retriever returns $\pi\!\sim\! q(\pi\mid g,\mathcal{K})$, and the gap between $\pi$ and $\pi^\star(g)$ is characterized by GUI drift: step insertion/removal, entry-point shift, control renaming, and goal mismatch. We denote by $\mathcal{D}_\Pi$ the (unknown, non-stationary) effective distribution over deployment-time priors.

\paragraph{Reliability-gated policy.} We introduce a latent reliability variable $r\!\in\!\mathcal{R}\!=\!\{\textsc{Follow},\textsc{Partial},\textsc{Ignore}\}$ and factorize
\begin{equation}
\label{eq:policy}
p_\theta(a_t\mid s_t,h_{<t},g,\pi)
=\sum_{r\in\mathcal{R}}\underbrace{p_\theta(r\mid s_t,h_{<t},g,\pi)}_{\text{reliability assessor}}\,\underbrace{p_\theta(a_t\mid s_t,h_{<t},g,\pi,r)}_{\text{reliability-conditioned actor}}.
\end{equation}
At inference, $(r,a_t)$ are realized jointly via autoregressive decoding, so Eq.~\ref{eq:policy} is implemented without explicit marginalization. This turns prior reliability from an unobserved nuisance into an \emph{explicit intermediate variable} the agent must commit to before acting.

\paragraph{Learning objective.} Given a task distribution $\mathcal{D}_{\text{task}}$ over $(g,h_{<t},s_t,a_t^\star)$ and a prior distribution $\mathcal{D}_\Pi(\cdot\mid g,h_{<t})$ whose reliability is determined by its relation to the trajectory (Sec.~\ref{sec:perturb}), we train
\begin{equation}
\label{eq:objective-generic}
\min_\theta\;
\mathbb{E}_{\substack{(g,h_{<t},s_t,a_t^\star)\sim\mathcal{D}_{\text{task}}\\ \pi\sim\mathcal{D}_\Pi}}\!
\Big[\!-\log p_\theta(r^\star\mid s_t,h_{<t},g,\pi)-\log p_\theta(a_t^\star\mid s_t,h_{<t},g,\pi,r^\star)\Big],
\end{equation}
where $r^\star$ is induced by the (known) perturbation producing $\pi$. Two design questions remain: how to obtain $\mathcal{D}_{\text{task}}$ at scale without annotation, and how to choose $\mathcal{D}_\Pi$ so that training-time priors cover deployment-time drift.

\subsection{Self-Explored Prior Acquisition}
\label{sec:exploration}

To supply application-specific priors without requiring human demonstrations, we explore real apps and organize observed transitions into a \emph{UI transition graph} $\mathcal{G}=(\mathcal{V},\mathcal{E})$, under three principles. Paths in $\mathcal{G}$ provide the operational experience from which we construct task-aligned priors and synthesizing tasks $\mathcal{D}_{\text{task}}$.

\paragraph{Semantic-region action abstraction.} Flat enumeration over raw controls wastes budget on repetitive list items and ads. A multimodal annotator instead partitions each screen into four region types, \emph{navigation}, \emph{functional}, \emph{homogeneous content}, \emph{distraction}, and the policy samples under a per-region budget $B_\text{nav}\!\geq\! B_\text{func}\!\gg\! B_\text{homog}\!\geq\! B_\text{distract}\!=\!0$. Each frontier node further maintains a candidate queue ordered by region priority, local exploration status, and control-level deduplication (by normalized bounds and semantic signatures), avoiding repeated interaction with semantically equivalent controls. The induced prior distribution concentrates on high-value functional paths likely to align with user intents (Cases in Appendix~\ref{app:case}).

\paragraph{Structure--visual state deduplication.} To prevent dynamic content (refresh, personalization, ads) from fragmenting the state space, two screens $s,s'$ are merged when
\begin{equation}
\label{eq:dedup}
S_{\text{xml}}(s,s')=\frac{|\mathcal{P}(s)\cap\mathcal{P}(s')|}{|\mathcal{P}(s)\cup\mathcal{P}(s')|}\geq\tau_{\text{xml}},\quad\text{and}\quad \phi_{\text{vis}}(s,s')=1,
\end{equation}
where $\mathcal{P}(\cdot)$ is the set of XML control paths and $\phi_\text{vis}$ is a VLM equivalence check. The conjunctive rule suppresses spurious nodes from recommendation refreshes while preventing false merges between visually similar but functionally distinct screens.

\paragraph{Recovery-aware backtracking as graph repair.} After each interaction, the explorer returns to the parent node. When a backtracking attempt lands on an unexpected screen, we treat it not as a failure but as evidence: the observed transition is recorded as a new edge, and we attempt graph-based replay from the root to the frontier, restarting the app only when no path exists. All edges in $\mathcal{G}$ are thus executable and verified by construction.

\paragraph{From graph to task-aligned priors.} We enumerate paths $\pi^\star=(e_1,\dots,e_L)$ in $\mathcal{G}$ and invoke a multimodal annotator to synthesize goals $g$ consistent with their functional endpoints. Each edge is described in natural language via its pre-state, action, and post-state to form $\bar e$, whose concatenation gives the trajectory-level prior $\pi^\star=[\bar e_1;\dots;\bar e_L]$. Unlike raw low-level logs, this abstraction exposes only action--outcome semantics and is invariant to coordinate perturbations or DOM reshuffles, making the prior a retrievable procedural hint that must still be verified at execution.

\subsection{Drift-Aware Perturbation as a Shaped Prior Distribution}
\label{sec:perturb}

Naive training on $\mathcal{D}_{\text{task}}\!\times\!\{\pi^\star\}$ induces unconditional prior trust and collapse under deployment drift. We instead construct $\mathcal{D}_\Pi$ using \emph{drift-aware perturbation operators} $T_k:\Pi\!\to\!\Pi$, $k\!\in\!\mathcal{K}\!=\!\{0,\dots,5\}$, grounded in Figure~\ref{fig:drift}(a)'s taxonomy, each with a deterministic rule assigning $r^\star$ (Table~\ref{tab:perturbation}). The training prior distribution is the mixture:
\begin{equation}
\label{eq:mixture}
\mathcal{D}_\Pi(\pi\mid \pi^\star)=\sum_{k\in\mathcal{K}} w_k\,\delta\!\big(\pi-T_k(\pi^\star)\big),\qquad \sum_k w_k=1,
\end{equation}
with $\{w_k\}$ as hyperparameters. Eq.~\ref{eq:mixture} (i) exposes the agent to all three reliability regimes in calibrated proportions, preventing degeneracy toward blind following ($w_0\!\to\!1$) or blind ignoring ($w_5\!\to\!1$), and (ii) provides an \emph{inductive bias} that $\mathrm{supp}(\mathcal{D}_\Pi)$ covers deployment-time drift, verified empirically in Sec.~\ref{sec:exp-robustness}.

\begin{table}[tbp]
\centering
\small
\renewcommand{\arraystretch}{1.0}
% Use the standard caption spacing configured above.
\caption{Drift-aware perturbation operators. $T_2$--$T_4$ additionally annotate trustworthy edge indices. $T_3,T_5$ draw from other paths in the same $\mathcal{G}$, keeping noisy priors \emph{in-distribution}.}
\begin{tabular}{@{}clllc@{}}
\toprule
$k$ & Operator $T_k$ & Empirical drift analog & Simulated failure mode & $r^\star$ \\
\midrule
$0$ & identity & clean retrieval & none & \textsc{Follow} \\
$1$ & surface rephrasing & control renaming & lexical/stylistic variation & \textsc{Follow} \\
$2$ & sub-path deletion & step removal & missing intermediate operation & \textsc{Partial} \\
$3$ & suffix branch replacement & entry-point/path shift & correct prefix, stale suffix & \textsc{Partial} \\
$4$ & irrelevant-op insertion & ads, permissions, overlays & noisy or unrelated step & \textsc{Partial} \\
$5$ & other-goal path substitution & wrong-goal retrieval & coherent but irrelevant prior & \textsc{Ignore} \\
\bottomrule
\end{tabular}
\label{tab:perturbation}
\end{table}

\paragraph{Coupling with exploration.} Because $T_3,T_5$ sample alternative paths from the same $\mathcal{G}$, exploration quality directly governs the \emph{hardness} of negative priors: a denser $\mathcal{G}$ yields \textsc{Partial}/\textsc{Ignore} negatives that are topologically adjacent to the target path, so scale and diversity of self-exploration are not merely data-quantity concerns but shape the hardness spectrum of $\mathcal{D}_\Pi$.

\subsection{Learning Objective and Training}
\label{sec:objective}

\paragraph{Step-level decomposition.} Each $(\pi^\star,g)$ yields a trajectory $\tau=(s_0,a_1,\dots,a_L,s_L)$, decomposed into $L{+}1$ step samples including a terminal \textsc{stop} supervision. For each step we sample $k\!\sim\!\mathrm{Cat}(w)$, apply $T_k$, and obtain an input $(s_t,h_{<t},g,\pi)$ with targets $(r^\star,a_t^\star)$.

\paragraph{Structured response and three consistency checks.} The model emits $\langle\texttt{Thought},\texttt{Action},\texttt{Tool}\rangle$, where \texttt{Thought} verbalizes the reliability decision, \texttt{Action} summarizes it in natural language, and \texttt{Tool} is the executable action. The reliability decision is structured around three checks: \emph{goal consistency} (does $\pi$ serve the same intent as $g$), \emph{history consistency} (is $h_{<t}$ compatible with a prefix of $\pi$), and \emph{UI grounding} (is the next suggested operation supported by $s_t$). Their joint outcome deterministically maps to $r$: all pass yields \textsc{Follow}; partial passes yield \textsc{Partial} with the set of trusted edge indices; a goal-consistency failure yields \textsc{Ignore}.

\paragraph{Teacher annotation.} \texttt{Thought} is supervised by a strong multimodal teacher conditioned on $(s_t,h_{<t},g,\pi,r^\star,a_t^\star)$, which we view as teacher-forced rationalization: since $r^\star$ and $a_t^\star$ are already fixed by the perturbation and the annotated trajectory, the teacher only articulates the justification. \texttt{Action} and \texttt{Tool} are reconstructed deterministically from $\pi^\star$, preventing hallucinated teacher actions from corrupting supervision.

\paragraph{Training loss.} Let $y=(\texttt{Thought},\texttt{Action},\texttt{Tool})$. We minimize the segment-weighted teacher-forced NLL
\begin{equation}
\label{eq:loss}
\mathcal{L}(\theta)=\mathbb{E}_{(s_t,h_{<t},g,\pi,y^\star)}\sum_{i}\lambda_{c(i)}\big(\!-\log p_\theta(y^\star_i\mid y^\star_{<i},s_t,h_{<t},g,\pi)\big),
\end{equation}
with $c(i)\!\in\!\{\texttt{thought},\texttt{action},\texttt{tool}\}$. Tokens encoding $r$ and the trusted-edge indices in \texttt{Thought} are upweighted, making Eq.~\ref{eq:loss} a concrete instantiation of Eq.~\ref{eq:objective-generic} with a rationale term. Since the same decision state is paired with priors of different reliability, the model cannot minimize Eq.~\ref{eq:loss} by copying $\pi$; it must learn when to follow, extract, or override.

\paragraph{Why explicit reliability supervision.} Explicit $r$ provides direct credit assignment over trustworthy prior segments and separates \textsc{Follow/Partial/Ignore} into distinct follow, filter, and override behaviors; its contribution is isolated empirically in Sec.~\ref{sec:ablation}.

\paragraph{Implementation.} We fine-tune Qwen3-VL and GUI-specialized VLM backbones with LoRA using Eq.~\ref{eq:loss}. Reliability assessment and action generation share one autoregressive decoder, requiring no auxiliary heads.

% Full hyperparameters, region-budget schedules, and perturbation-weight ablations are in Appendix~\ref{app:training-evaluation}.

\begin{figure}[tbp]
    \centering
    \begin{subfigure}[t]{0.28\textwidth}
        \centering
        \includegraphics[width=\linewidth,height=2.5cm,keepaspectratio]{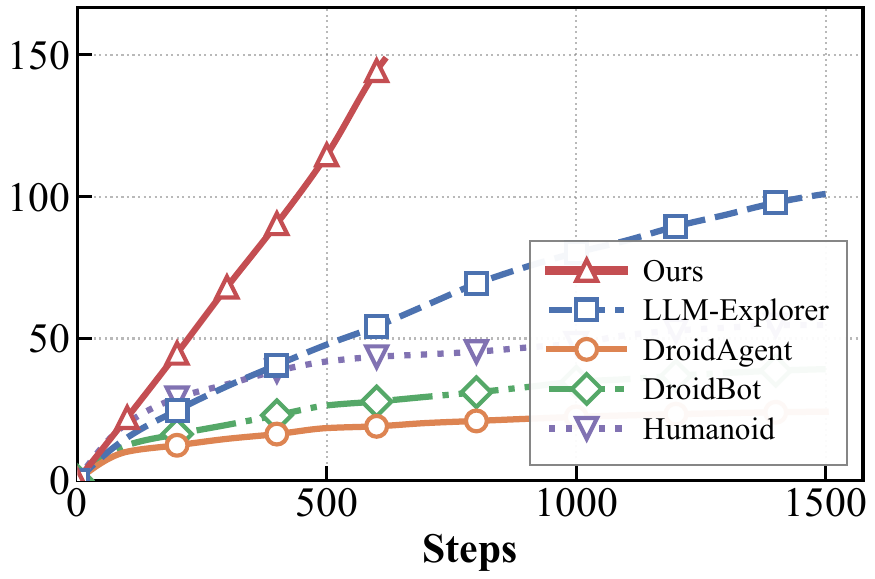}
        \caption{\# of Unique Screens}
        \label{fig:first}
    \end{subfigure}
    \hfill
    \begin{subfigure}[t]{0.28\textwidth}
        \centering
        \includegraphics[width=\linewidth,height=2.5cm,keepaspectratio]{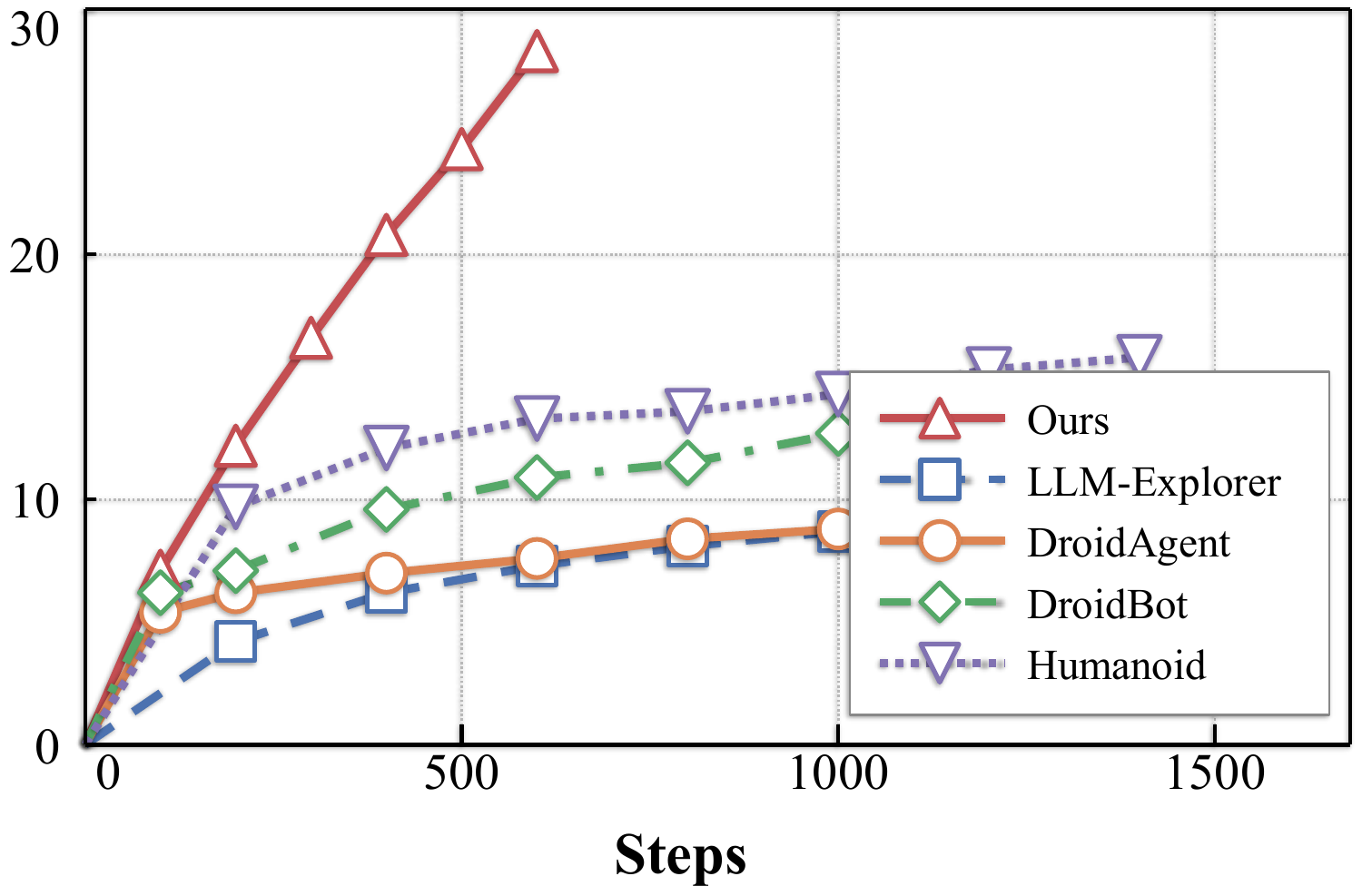}
        \caption{\# of Explored Activities}
        \label{fig:second}
    \end{subfigure}
    \hfill
    \begin{subfigure}[t]{0.42\textwidth}
        \centering
        \includegraphics[width=\linewidth,height=2.5cm,keepaspectratio]{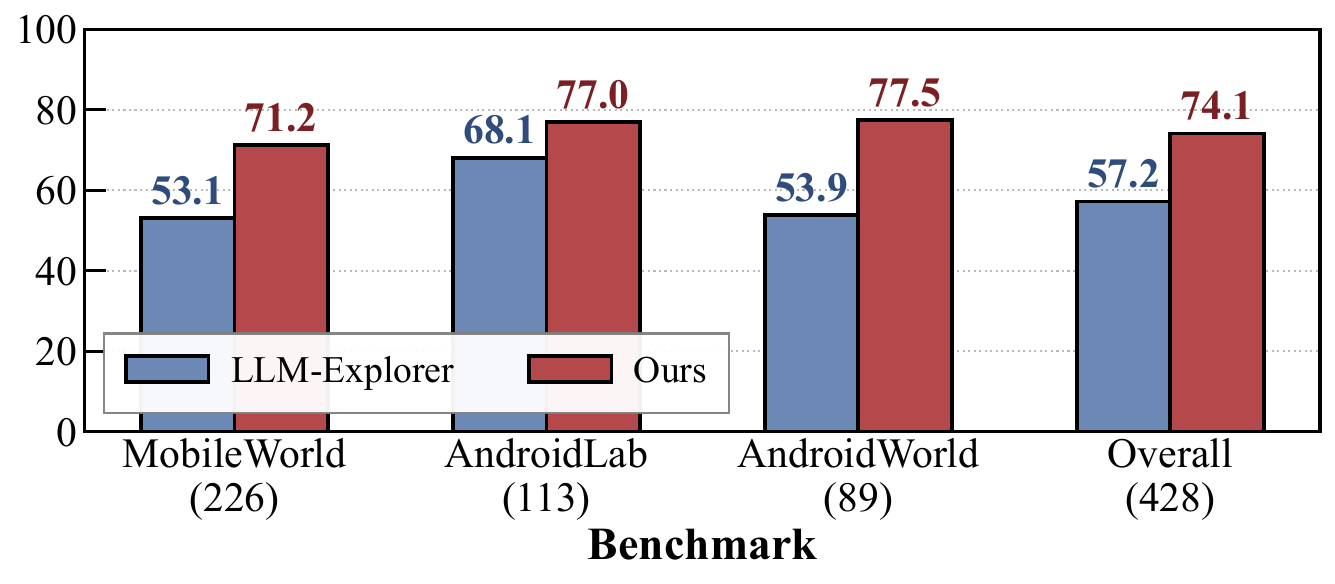}
        \caption{Subtask Coverage}
        \label{fig:third}
    \end{subfigure}
    \caption{Exploration efficiency and downstream utility. (a) \# of average deduplicated screens vs.\ steps. (b) \# of average activities. (c) Subtask coverage (\%) on MobileWorld / AndroidLab / AndroidWorld under the same budget; numbers in parentheses denote subtask counts.}
    \label{fig:exploration}
\end{figure}

\section{Experiments}

We evaluate whether \textsc{SAGE} effectively couples knowledge acquisition with robust utilization of imperfect priors. Specifically, we examine (i) whether self-exploration yields denser and more task-relevant priors, and (ii) whether reliability-gated policies act robustly under imperfect priors. Learning experiments draw from three trajectory sources in the same format: \textbf{CUTG} (ours, from self-explored graphs) and two public sets \textbf{CMGUI} and \textbf{ChiM-Nav}; full statistics are in Appendix~\ref{app:dataset}.

\subsection{Exploration Efficiency and Task-Relevant Coverage}
\label{sec:exp-explore}

\paragraph{Setup.}
We evaluate exploration along two complementary axes.

\emph{(i) In-the-wild efficiency.} On 20 Android apps spanning 5 categories (social, shopping, travel, media, utilities; full list in Appendix~\ref{app: Subset}), we compare against four representative baselines: two LLM-driven explorers, LLM-Explorer~\citep{llm_explorer} and DroidAgent~\citep{droidagent}, and two classical UI explorers, DroidBot~\citep{droidbot} and Humanoid~\citep{li2019humanoid}. All methods start from identical initial states (freshly installed app, logged-out where applicable). For fair comparison, all LLM-based methods use the same backbone (Qwen3-VL-Plus) and the same screen-deduplication rule based on view-hierarchy hashing; classical baselines follow their original implementations. We report two efficiency curves: \# of \emph{deduplicated screens} and \# of \emph{distinct activities} reached as a function of step count.\looseness=-1

\emph{(ii) Task-relevant coverage.} To measure whether exploration actually reaches interfaces that matter for downstream tasks, we evaluate on three benchmarks, MobileWorld, AndroidLab, and AndroidWorld (129 / 85 / 75 tasks). Each task is decomposed into atomic subtasks (226 / 113 / 89 in total), and each subtask is tied to a single \emph{target interface} that must be visited to complete it. A subtask is \emph{covered} if any explored trajectory visits its target interface (matched by activity name and a semantic check on key UI elements). The decomposition and target-interface labels were independently produced by two of the authors, and disagreements were resolved through discussion, ensuring consistency with the ground truth provided by the benchmarks. We compare against the strongest in-the-wild baseline (LLM-Explorer) under the same per-app budget, and report per-benchmark coverage rates as well as the overall average.\looseness=-1

\paragraph{Results.}
Figure~\ref{fig:exploration}(a,b) shows that our explorer reaches $\sim$148 deduplicated screens and $\sim$28 activities within 600 steps, whereas LLM-Explorer requires $\sim$1{,}500 steps to reach 101 screens and 10 activities, and the remaining baselines plateau below 55 screens. On benchmark coverage (Figure~\ref{fig:exploration}(c)), our method improves over LLM-Explorer on every benchmark (71.2 vs.\ 53.1, 77.0 vs.\ 68.1, 77.5 vs.\ 53.9), with an overall gain of $+$16.9 points (74.1 vs.\ 57.2). Together, the two views suggest that \textsc{SAGE} covers more interfaces required for the tasks, producing a denser pool of candidate priors.

% llm-explorer 效果一般，因为虚拟环境运行慢

\subsection{Robustness to Imperfect Priors}
\label{sec:exp-robustness}

\begin{table}[htbp]
\centering
\footnotesize
\setlength{\tabcolsep}{3.2pt}
\renewcommand{\arraystretch}{1.0}
\caption{Accuracy gains (in percentage points) over the no-prior baseline across
different prior conditions. Higher is better. Shaded rows denote our method;
subscripts indicate absolute improvement over the SFT baseline (\up{}\,/\,\dn{}). 
Avg.\ is the unweighted mean over all prior conditions.}
\label{tab:main-results-ours}
\begin{tabular}{@{}l l c c c c c c c@{}}
\toprule
\multirow{2}{*}{\textbf{Backbone}}
 & \multirow{2}{*}{\textbf{Method}}
 & \multicolumn{7}{c}{
     \textbf{Accuracy gain over no-prior (pp)} \;\; $\uparrow$
   } \\
\cmidrule(lr){3-9}
 & & \textbf{Correct} & \textbf{Local} & \textbf{Noise}
   & \textbf{Omit} & \textbf{Rewrite} & \textbf{Other}
   & \textbf{Avg.} \\
\midrule

% ================= Qwen3-VL-2B =================
\multirow{2}{*}{Qwen3-VL-2B}
  & SFT
  & 6.53 & 0.41 & 6.53 & 0.82 & 3.27 & 0.00 & 2.93 \\
  & \cellcolor{rowshade}\textsc{Ours}
  & \cellcolor{rowshade}\textbf{21.63}\,\up{15.10}
  & \cellcolor{rowshade}\textbf{4.90}\,\up{4.49}
  & \cellcolor{rowshade}\textbf{13.06}\,\up{6.53}
  & \cellcolor{rowshade}\textbf{9.80}\,\up{8.98}
  & \cellcolor{rowshade}\textbf{11.84}\,\up{8.57}
  & \cellcolor{rowshade}\textbf{8.98}\,\up{8.98}
  & \cellcolor{rowshade}\textbf{11.70}\,\up{8.77} \\
\cmidrule(lr){1-9}

% ================= Qwen3-VL-4B =================
\multirow{2}{*}{Qwen3-VL-4B}
  & SFT
  & 18.37 & 3.67 & 11.02 & 4.49 & 8.57 & 4.49 & 8.44 \\
  & \cellcolor{rowshade}\textsc{Ours}
  & \cellcolor{rowshade}\textbf{21.63}\,\up{3.26}
  & \cellcolor{rowshade}\textbf{6.53}\,\up{2.86}
  & \cellcolor{rowshade}\textbf{13.47}\,\up{2.45}
  & \cellcolor{rowshade}\textbf{7.35}\,\up{2.86}
  & \cellcolor{rowshade}\textbf{13.47}\,\up{4.90}
  & \cellcolor{rowshade}\textbf{4.90}\,\up{0.41}
  & \cellcolor{rowshade}\textbf{11.23}\,\up{2.79} \\
\cmidrule(lr){1-9}

% ================= Qwen3-VL-8B =================
\multirow{2}{*}{Qwen3-VL-8B}
  & SFT
  & 13.06 & -3.67 & 3.67 & -1.63 & 4.90 & -1.63 & 2.45 \\
  & \cellcolor{rowshade}\textsc{Ours}
  & \cellcolor{rowshade}\textbf{20.41}\,\up{7.35}
  & \cellcolor{rowshade}\textbf{8.16}\,\up{11.83}
  & \cellcolor{rowshade}\textbf{11.84}\,\up{8.17}
  & \cellcolor{rowshade}\textbf{8.98}\,\up{10.61}
  & \cellcolor{rowshade}\textbf{15.51}\,\up{10.61}
  & \cellcolor{rowshade}\textbf{6.12}\,\up{7.75}
  & \cellcolor{rowshade}\textbf{11.84}\,\up{9.39} \\
\cmidrule(lr){1-9}

% ================= MAI-UI-2B =================
\multirow{2}{*}{MAI-UI-2B}
  & SFT
  & 10.61 & 0.41 & 6.12 & 3.27 & 4.08 & -1.63 & 3.81 \\
  & \cellcolor{rowshade}\textsc{Ours}
  & \cellcolor{rowshade}\textbf{17.14}\,\up{6.53}
  & \cellcolor{rowshade}\textbf{2.04}\,\up{1.63}
  & \cellcolor{rowshade}\textbf{8.57}\,\up{2.45}
  & \cellcolor{rowshade}\textbf{5.31}\,\up{2.04}
  & \cellcolor{rowshade}\textbf{8.98}\,\up{4.90}
  & \cellcolor{rowshade}\textbf{0.41}\,\up{2.04}
  & \cellcolor{rowshade}\textbf{7.08}\,\up{3.27} \\
\cmidrule(lr){1-9}

% ================= GELab-Zero-4B =================
\multirow{2}{*}{GELab-Zero-4B}
  & SFT
  & 6.53 & -0.41 & 2.45 & 2.04 & 2.45 & -3.27 & 1.63 \\
  & \cellcolor{rowshade}\textsc{Ours}
  & \cellcolor{rowshade}\textbf{12.24}\,\up{5.71}
  & \cellcolor{rowshade}\textbf{4.49}\,\up{4.90}
  & \cellcolor{rowshade}\textbf{7.35}\,\up{4.90}
  & \cellcolor{rowshade}\textbf{4.49}\,\up{2.45}
  & \cellcolor{rowshade}\textbf{8.98}\,\up{6.53}
  & \cellcolor{rowshade}\textbf{4.90}\,\up{8.17}
  & \cellcolor{rowshade}\textbf{7.08}\,\up{5.45} \\
\bottomrule
\end{tabular}
\end{table}

\paragraph{Setup.}
Using self-explored CUTG trajectories, we test whether reliability-gated training improves prior utilization under controlled imperfections. \textbf{SFT} is trained with standard action supervision and receives priors only at inference, whereas \textbf{Ours} uses prior-conditioned Full-CoT supervision with an explicit {\textsc{FOLLOW}, \textsc{PARTIAL}, \textsc{IGNORE}} tag; both are evaluated with the same priors. Thus, the SFT comparison evaluates the benefit of the full prior-aware training scheme over direct prior injection; the contribution of explicit reliability supervision is isolated separately against Part-CoT in Sec.~\ref{sec:ablation}.\looseness=-1

\paragraph{Evaluation protocol.}
Each test instance is presented under six prior conditions, each with a ground-truth reliability tag that a well-calibrated policy should emit:
\textbf{Correct} (\textsc{follow}),
\textbf{Rewrite} (paraphrased; \textsc{follow}),
\textbf{Omit} (intermediate steps removed; \textsc{partial}),
\textbf{Noise} (irrelevant steps inserted; \textsc{partial}),
\textbf{Local} (locally perturbed suffix; \textsc{partial}),
and \textbf{Other} (path toward a different goal; \textsc{ignore}).
We report the marginal step-accuracy gain
\(\Delta \;=\; \mathrm{Acc}(\text{with prior}) - \mathrm{Acc}(\text{no prior}) \).
A robust agent should show large positive $\Delta$ under \textsc{follow} conditions and non-negative $\Delta$ under \textsc{partial}/\textsc{ignore} conditions.\looseness=-1

\paragraph{Results.}
% 注: 缩减篇幅改短
% Table~\ref{tab:main-results-ours} reports $\Delta$ on CUTG across five backbones; three patterns stand out.
% \emph{(i) Consistent average gains.} -Ours improves the averaged $\Delta$ on every backbone, by $+8.77$, $+2.79$, $+9.39$, $+3.27$, and $+5.45$\,pp on Qwen3-VL-\{2B,4B,8B\}, MAI-UI-2B, and GELab-Zero-4B; on the Qwen3-VL family, -Ours reaches $11.23$--$11.84$\,pp vs.\ $2.45$--$8.44$\,pp for -SFT.
% \emph{(ii) Robustness under imperfect priors.} -SFT turns negative on Qwen3-VL-8B under Local ($-3.67$), Omit ($-1.63$), and Other ($-1.63$), showing that naive retrieval augmentation can hurt; -Ours removes these regressions and reaches $+8.16$, $+8.98$, and $+6.12$\,pp on the same conditions.
% \emph{(iii) No conservatism under reliable priors.} The Correct-prior gain is preserved or amplified ($6.53\!\rightarrow\!21.63$ on Qwen3-VL-2B, $13.06\!\rightarrow\!20.41$ on Qwen3-VL-8B), so explicit gating does not make the model ignore good priors. In short, learning to \emph{judge} a prior before using it strictly dominates action-only supervision in the reliable-vs.-misleading trade-off.\looseness=-1
Table~\ref{tab:main-results-ours} shows that Ours yields higher
prior-induced gains than SFT across all five backbones and six prior
conditions, improving the backbone-level means by $2.79$--$9.39$\,pp.
On Qwen3-VL-8B, SFT incurs negative gains under Local, Omit, and Other
($-3.67$, $-1.63$, and $-1.63$\,pp), whereas Ours achieves
$8.16$, $8.98$, and $6.12$\,pp.
Ours also obtains larger gains from Correct priors on every backbone,
including $6.53\!\rightarrow\!21.63$\,pp on Qwen3-VL-2B.
Thus, the full training scheme improves utilization of both reliable and imperfect priors in the evaluated settings. To verify that reliability assessment extends beyond synthetic perturbations, we further evaluate on manually annotated held-out real cross-version drift; Appendix~\ref{app:real-drift-reliability} reports class-wise F1 and Macro-F1 across four backbones.

\subsection{Online Benchmark}
\label{sec:exp-onlinebench}

\begin{table}[htbp]
\centering
\footnotesize
\setlength{\tabcolsep}{7pt}
\renewcommand{\arraystretch}{1.0}
\caption{
Mean online task success rate gains
($\Delta$, in percentage points) over each method's own no-prior
baseline on 144 tasks across three benchmarks.
Values are averaged equally over all prior conditions.
Higher is better. Shaded rows denote our method;
subscripts indicate absolute improvement over the SFT baseline (\up{}\,/\,\dn{}).
}
\label{tab:online-results}

\begin{tabular}{@{}l l c c c@{}}
\toprule
\multirow{2}{*}{\textbf{Backbone}}
  & \multirow{2}{*}{\textbf{Method}}
  & \multicolumn{3}{c}{
      \textbf{Mean success rate gain over no-prior (pp)}
      \;\; $\uparrow$
    } \\
\cmidrule(lr){3-5}
  & & \textbf{AndroidWorld}
    & \textbf{AndroidLab}
    & \textbf{MobileWorld} \\
\midrule

% ================= Qwen3-VL-4B =================
\multirow{2}{*}{Qwen3-VL-4B}
  & SFT
  & -6.80
  & -10.32
  & -6.20 \\
  & \cellcolor{rowshade}\textsc{Ours}
  & \cellcolor{rowshade}\textbf{-2.60}\,\up{4.20}
  & \cellcolor{rowshade}\textbf{0.31}\,\up{10.63}
  & \cellcolor{rowshade}\textbf{7.70}\,\up{13.90} \\
\cmidrule(lr){1-5}

% ================= GUI-OWL-1.5-4B =================
\multirow{2}{*}{GUI-OWL-1.5-4B}
  & SFT
  & -7.66
  & -1.52
  & -2.34 \\
  & \cellcolor{rowshade}\textsc{Ours}
  & \cellcolor{rowshade}\textbf{-1.70}\,\up{5.96}
  & \cellcolor{rowshade}\textbf{1.26}\,\up{2.78}
  & \cellcolor{rowshade}\textbf{5.22}\,\up{7.56} \\

\bottomrule
\end{tabular}
\end{table}

\paragraph{Setup.} We compare \textbf{SFT} and \textbf{Ours} on online tasks with exploration-derived priors, using Qwen3-VL-4B and GUI-OWL-1.5-4B trained on CUTG under the offline protocol. We evaluate 144 exploration-covered tasks from AndroidWorld, AndroidLab, and MobileWorld, with task-specific priors constructed from the exploration graphs. Execution environments are initialized according to task configurations rather than exploration states, so even graph-relative \textit{Correct} priors may not fully match the execution tasks.

\paragraph{Results.} Table~\ref{tab:online-results} shows that \textbf{Ours} achieves higher mean $\Delta$ than \textbf{SFT} for both backbones on all three benchmarks, with improvements of $2.78$--$13.90$ pp. On AndroidWorld, where priors hurt both methods, \textbf{Ours} reduces the declines from $-6.80$ to $-2.60$ pp and from $-7.66$ to $-1.70$ pp. These results support the effectiveness of reliability-aware training in online prior utilization, while showing that imperfect priors can still impair task completion.

\subsection{Cross-Dataset Generalization}
\label{sec:transfer}

\begin{table*}[!t]
\centering
\caption{Cross-dataset generalization under different prior conditions.
Values are accuracy gains (in percentage points) over the no-prior setting.
Each pair of columns reports \textbf{Ours} (shaded) and \textbf{SFT};
the larger value in each pair is \textbf{bolded}.}
\label{tab:cross-dataset-detailed}
\setlength{\tabcolsep}{3.2pt}
\renewcommand{\arraystretch}{1.1}
\footnotesize
% --- helper macros: now each one expands into TWO cells ---
\newcommand{\bU}[2]{\textbf{#1} & #2}          % Ours bold
\newcommand{\bS}[2]{#1 & \textbf{#2}}          % SFT  bold
\newcommand{\bB}[2]{\textbf{#1} & \textbf{#2}} % both bold
\resizebox{\textwidth}{!}{%
\begin{tabular}{@{}l l *{7}{>{\columncolor{gray!12}}c c}@{}}
\toprule
\multirow{2}{*}{\textbf{Model}} & \multirow{2}{*}{\textbf{Train\,$\rightarrow$\,Test}}
 & \multicolumn{2}{c}{\textbf{Corr.}}
 & \multicolumn{2}{c}{\textbf{Local}}
 & \multicolumn{2}{c}{\textbf{Noise}}
 & \multicolumn{2}{c}{\textbf{Other}}
 & \multicolumn{2}{c}{\textbf{Rewrite}}
 & \multicolumn{2}{c}{\textbf{Omit}}
 & \multicolumn{2}{c}{\textbf{Avg.}} \\
\cmidrule(lr){3-4}\cmidrule(lr){5-6}\cmidrule(lr){7-8}\cmidrule(lr){9-10}%
\cmidrule(lr){11-12}\cmidrule(lr){13-14}\cmidrule(lr){15-16}
 & & \textbf{Ours} & SFT & \textbf{Ours} & SFT & \textbf{Ours} & SFT
   & \textbf{Ours} & SFT & \textbf{Ours} & SFT & \textbf{Ours} & SFT
   & \textbf{Ours} & SFT \\
\midrule
% ============== Qwen3-2B ==============
\multirow{6}{*}{Qwen3-2B}
 & CMGUI\,$\rightarrow$\,CUTG     & \bU{13.06}{6.94}  & \bU{6.53}{-3.27}  & \bU{7.35}{4.08}   & \bU{2.86}{-4.08}  & \bU{4.90}{2.45}   & \bU{4.90}{0.41}   & \bU{6.60}{1.09}  \\
 & CMGUI\,$\rightarrow$\,ChiM & \bU{15.50}{3.10}  & \bU{12.40}{0.00}  & \bU{15.50}{3.88}  & \bU{1.55}{-4.65}  & \bU{13.95}{3.10}  & \bU{13.18}{5.43}  & \bU{12.01}{1.81} \\
 & CUTG\,$\rightarrow$\,CMGUI     & \bU{7.19}{3.77}   & \bU{3.77}{3.42}   & \bU{6.85}{5.48}   & \bS{-4.79}{-3.42} & \bU{3.42}{2.40}   & \bU{7.53}{6.16}   & \bU{3.99}{2.97}  \\
 & CUTG\,$\rightarrow$\,ChiM  & \bU{10.85}{2.33}  & \bU{8.53}{0.00}   & \bU{6.98}{2.33}   & \bS{-1.55}{-1.55} & \bU{7.75}{3.88}   & \bU{10.85}{7.75}  & \bU{7.23}{2.46}  \\
 & ChiM\,$\rightarrow$\,CMGUI & \bU{6.51}{0.00}   & \bU{5.48}{-1.71}  & \bU{4.79}{1.71}   & \bU{-0.34}{-2.40} & \bU{5.48}{3.77}   & \bU{7.88}{2.05}   & \bU{4.97}{0.57}  \\
 & ChiM\,$\rightarrow$\,CUTG  & \bU{8.98}{5.31}   & \bU{0.41}{-2.45}  & \bU{7.76}{4.49}   & \bU{2.04}{-1.22}  & \bU{4.49}{1.22}   & \bU{4.49}{-0.82}  & \bU{4.70}{1.09}  \\
\midrule
% ============== Qwen3-4B ==============
\multirow{6}{*}{Qwen3-4B}
 & CMGUI\,$\rightarrow$\,CUTG     & \bU{13.47}{13.06} & \bU{6.53}{2.45}   & \bU{9.39}{6.94}   & \bU{5.71}{1.63}   & \bS{3.67}{4.08}   & \bU{4.49}{3.27}   & \bU{7.21}{5.24}  \\
 & CMGUI\,$\rightarrow$\,ChiM & \bU{12.40}{6.98}  & \bU{8.53}{4.65}   & \bU{13.95}{5.43}  & \bU{3.88}{-1.55}  & \bU{11.63}{7.75}  & \bU{7.75}{6.20}   & \bU{9.69}{4.91}  \\
 & CUTG\,$\rightarrow$\,CMGUI     & \bU{7.88}{3.42}   & \bU{5.82}{1.37}   & \bU{5.82}{1.71}   & \bU{-0.34}{-1.03} & \bU{5.48}{4.45}   & \bU{7.88}{4.79}   & \bU{5.42}{2.45}  \\
 & CUTG\,$\rightarrow$\,ChiM  & \bU{16.28}{13.18} & \bU{10.85}{4.65}  & \bU{14.73}{12.40} & \bS{1.55}{4.65}   & \bS{8.53}{10.08}  & \bU{13.95}{13.95} & \bU{10.98}{9.82} \\
 & ChiM\,$\rightarrow$\,CMGUI & \bU{3.42}{1.37}   & \bU{2.40}{1.37}   & \bU{3.08}{1.37}   & \bU{0.68}{-1.37}  & \bU{2.40}{1.03}   & \bU{3.42}{1.71}   & \bU{2.57}{0.91}  \\
 & ChiM\,$\rightarrow$\,CUTG  & \bU{17.96}{11.43} & \bU{6.94}{0.82}   & \bU{11.84}{4.90}  & \bU{4.90}{-0.41}  & \bU{8.98}{3.67}   & \bU{5.71}{2.86}   & \bU{9.39}{3.88}  \\
\midrule
% ============== Qwen3-8B ==============
\multirow{6}{*}{Qwen3-8B}
 & CMGUI\,$\rightarrow$\,CUTG     & \bS{14.69}{15.51} & \bU{6.12}{3.67}   & \bU{12.24}{11.84} & \bU{5.71}{2.86}   & \bS{6.53}{6.94}   & \bU{4.49}{4.49}   & \bU{8.30}{7.55}  \\
 & CMGUI\,$\rightarrow$\,ChiM & \bU{7.75}{2.33}   & \bU{4.65}{1.55}   & \bU{8.53}{2.33}   & \bU{-1.55}{-5.43} & \bU{5.43}{3.10}   & \bU{8.53}{5.43}   & \bU{5.56}{1.55}  \\
 & CUTG\,$\rightarrow$\,CMGUI     & \bU{12.67}{5.14}  & \bU{9.59}{4.45}   & \bU{12.67}{3.77}  & \bS{-0.68}{0.68}  & \bU{8.22}{3.42}   & \bU{11.30}{4.79}  & \bU{8.96}{3.71}  \\
 & CUTG\,$\rightarrow$\,ChiM  & \bU{10.85}{10.08} & \bU{7.75}{6.98}   & \bS{10.08}{11.63} & \bS{-1.55}{3.10}  & \bU{13.95}{9.30}  & \bU{11.63}{10.08} & \bU{8.79}{8.53}  \\
 & ChiM\,$\rightarrow$\,CMGUI & \bU{2.74}{2.05}   & \bU{2.40}{1.37}   & \bU{2.05}{1.37}   & \bU{1.71}{-0.68}  & \bU{2.05}{1.71}   & \bU{2.74}{2.05}   & \bU{2.28}{1.31}  \\
 & ChiM\,$\rightarrow$\,CUTG  & \bU{16.33}{10.61} & \bU{4.90}{1.63}   & \bU{11.43}{8.98}  & \bU{5.31}{2.86}   & \bU{11.02}{6.94}  & \bU{3.67}{3.27}   & \bU{8.78}{5.71}  \\
\midrule
% ============== MAI-2B ==============
\multirow{6}{*}{MAI-2B}
 & CMGUI\,$\rightarrow$\,CUTG     & \bS{12.65}{13.06} & \bU{4.49}{0.00}   & \bS{6.94}{9.39}   & \bS{1.63}{3.67}   & \bS{6.12}{8.98}   & \bS{3.27}{6.12}   & \bS{5.85}{6.87}  \\
 & CMGUI\,$\rightarrow$\,ChiM & \bU{20.16}{9.30}  & \bU{13.18}{0.78}  & \bU{15.50}{8.53}  & \bU{0.78}{-4.65}  & \bU{12.40}{6.98}  & \bU{13.95}{10.08} & \bU{12.66}{5.17} \\
 & CUTG\,$\rightarrow$\,CMGUI     & \bU{12.33}{6.51}  & \bU{4.79}{3.77}   & \bU{7.88}{6.16}   & \bU{3.42}{-8.90}  & \bU{9.25}{7.19}   & \bU{11.64}{6.51}  & \bU{8.22}{3.54}  \\
 & CUTG\,$\rightarrow$\,ChiM  & \bU{10.08}{10.08} & \bU{7.75}{4.65}   & \bU{10.85}{10.85} & \bU{-0.78}{-3.10} & \bS{6.98}{9.30}   & \bS{11.63}{12.40} & \bU{7.75}{7.36}  \\
 & ChiM\,$\rightarrow$\,CMGUI & \bU{5.14}{4.79}   & \bU{1.71}{1.71}   & \bS{3.77}{5.82}   & \bU{-3.42}{-6.85} & \bS{3.42}{5.14}   & \bU{4.79}{4.79}   & \bB{2.57}{2.57}  \\
 & ChiM\,$\rightarrow$\,CUTG  & \bU{17.14}{10.61} & \bU{1.63}{-2.86}  & \bU{8.57}{6.53}   & \bU{3.67}{0.82}   & \bU{8.98}{4.90}   & \bU{6.12}{3.27}   & \bU{7.68}{3.88}  \\
\midrule
% ============== GELab-4B ==============
\multirow{6}{*}{GELab-4B}
 & CMGUI\,$\rightarrow$\,CUTG     & \bU{13.06}{-2.45} & \bU{5.71}{-2.86}  & \bU{8.57}{-2.45}  & \bU{4.08}{-4.49}  & \bU{6.12}{-4.08}  & \bU{2.45}{-8.16}  & \bU{6.67}{-4.08} \\
 & CMGUI\,$\rightarrow$\,ChiM & \bU{10.08}{4.65}  & \bU{6.20}{4.65}   & \bU{10.85}{7.75}  & \bS{-0.78}{3.88}  & \bU{8.53}{3.10}   & \bU{10.08}{6.98}  & \bU{7.49}{5.17}  \\
 & CUTG\,$\rightarrow$\,CMGUI     & \bU{5.14}{-3.08}  & \bU{2.40}{-3.08}  & \bU{4.45}{-2.40}  & \bU{-0.68}{-4.11} & \bU{6.51}{-1.03}  & \bU{5.82}{-1.37}  & \bU{3.94}{-2.51} \\
 & CUTG\,$\rightarrow$\,ChiM  & \bU{10.85}{-2.33} & \bU{6.20}{-6.98}  & \bU{8.53}{-6.98}  & \bU{-4.65}{-5.43} & \bU{1.55}{-4.65}  & \bU{7.75}{1.55}   & \bU{5.04}{-4.14} \\
 & ChiM\,$\rightarrow$\,CMGUI & \bU{4.45}{3.77}   & \bU{4.45}{2.05}   & \bU{6.51}{3.08}   & \bU{3.42}{-1.03}  & \bS{0.34}{3.77}   & \bU{5.48}{3.08}   & \bU{4.11}{2.45}  \\
 & ChiM\,$\rightarrow$\,CUTG  & \bU{10.61}{9.39}  & \bU{1.63}{0.00}   & \bU{7.76}{6.53}   & \bS{1.63}{2.04}   & \bU{6.94}{3.27}   & \bS{1.63}{2.45}   & \bU{5.03}{3.95}  \\
\bottomrule
\end{tabular}}
\end{table*}

\paragraph{Setup.} To test whether prior assessment transfers as a skill rather than as dataset-specific memorization, we evaluate all six train$\to$test pairs over \{CUTG, CMGUI, ChiM-Nav\} (in-domain diagonals excluded) across five backbones, retaining the six-way perturbation taxonomy of Sec~\ref{sec:exp-robustness}. Each cell of Table~\ref{tab:cross-dataset-detailed} reports Ours\,/\,SFT as percentage-point gains over the no-prior baseline.

% 注: 缩减篇幅改短
% \paragraph{Results.} Three patterns emerge consistently across backbones. \emph{(i) Higher average transfer.} Averaged over all off-diagonal pairs, Ours lifts the mean gain from $3.19$ to $6.81$\,pp and matches or exceeds SFT on the Avg.\ column in $29/30$ settings (strictly better in $28$). \emph{(ii) Gains concentrate on reliable and partially reliable priors.} Under Corr., Local, Noise, and Omit, SFT frequently collapses to zero or negative transfer---e.g., GELab-4B trained on CMGUI or CUTG yields a negative SFT mean on unseen domains while Ours stays positive---which is consistent with explicit assessment extracting more signal from usable priors and suppressing noise under distribution shift. \emph{(iii) Misleading priors remain the hardest regime.} Under the \emph{Other} condition both methods occasionally degrade and a few cells favor SFT, suggesting that rejecting goal-mismatched priors is the transfer mode that benefits least from our mechanism. Taken together, these off-diagonal results suggest that prior assessment generalizes as a \emph{mechanism} for selectively using GUI priors, rather than as memorization of dataset-specific paths.\looseness=-1
\paragraph{Results.}
Across all 30 backbone--transfer settings, Ours raises the mean prior-induced gain from $3.19$ to $6.81$\,pp, exceeding SFT on mean $\Delta$ in $28$ settings and matching it in one (Table~\ref{tab:cross-dataset-detailed}). For GELab-4B trained on CUTG, SFT has negative mean gains on both unseen datasets, whereas Ours remains positive. Goal-mismatched priors remain challenging: under Other, both methods sometimes degrade, and SFT performs better in some settings. These results support transfer of prior-assessment behavior across the evaluated datasets, while indicating limited robustness to misleading priors.\looseness=-1

\subsection{Ablation Study}
\label{sec:ablation}

\begin{table}[htbp]
\centering
\setlength{\tabcolsep}{1.5pt}
\renewcommand{\arraystretch}{1.0}
\small
% 注: 缩减篇幅改短
% \caption{In-domain ablation under diagonal train--test settings.
% Values are average accuracy gains (in percentage points) over each
% method's own no-prior baseline across all prior conditions.
% \textbf{Ori.} denotes the original base model;
% \textbf{SFT} denotes direct fine-tuning;
% \textbf{Part-CoT} retains CoT supervision without explicit
% reliability-label supervision; and \textbf{Ours} denotes Full-CoT.
% Average is the unweighted mean across the three datasets.
% Ours is shaded, and the highest value within each group is bolded.}
\caption{In-domain ablation.
Values are mean accuracy gains (pp) over each variant's own
no-prior baseline, averaged equally across prior conditions.
Average further averages over the three datasets.
Ours is shaded; the best value in each group is bolded.}
\label{tab:ablation}
\resizebox{\textwidth}{!}{%
\begin{tabular}{@{}l *{4}{ccc>{\columncolor{gray!12}}c}@{}}
\toprule
\multicolumn{1}{c}{\multirow{2}{*}[-2.0ex]{\textbf{Model}}}
 & \multicolumn{4}{c}{\textbf{CUTG}}
 & \multicolumn{4}{c}{\textbf{CMGUI}}
 & \multicolumn{4}{c}{\textbf{ChiM-Nav}}
 & \multicolumn{4}{c}{\textbf{Average}} \\
\cmidrule(lr){2-5}
\cmidrule(lr){6-9}
\cmidrule(lr){10-13}
\cmidrule(lr){14-17}
 & Ori. & SFT
 & \begin{tabular}[c]{@{}c@{}}Part-\\CoT\end{tabular}
 & Ours
 & Ori. & SFT
 & \begin{tabular}[c]{@{}c@{}}Part-\\CoT\end{tabular}
 & Ours
 & Ori. & SFT
 & \begin{tabular}[c]{@{}c@{}}Part-\\CoT\end{tabular}
 & Ours
 & Ori. & SFT
 & \begin{tabular}[c]{@{}c@{}}Part-\\CoT\end{tabular}
 & Ours \\
\midrule
Qwen3-VL-2B
 & 3.06 & 2.93 & 6.53 & \textbf{11.70}
 & 1.03 & 3.66 & 2.17 & \textbf{4.97}
 & 4.39 & 1.03 & 8.66 & \textbf{9.82}
 & 2.83 & 2.54 & 5.79 & \textbf{8.83} \\
Qwen3-VL-4B
 & 4.83 & 8.44 & 4.42 & \textbf{11.23}
 & 0.57 & 1.49 & 2.28 & \textbf{2.40}
 & 6.59 & 6.72 & 5.95 & \textbf{10.21}
 & 4.00 & 5.55 & 4.22 & \textbf{7.95} \\
Qwen3-VL-8B
 & 5.31 & 2.45 & 9.59 & \textbf{11.84}
 & 0.57 & 2.05 & \textbf{2.40} & 1.77
 & 4.00 & 5.82 & \textbf{11.11} & 10.59
 & 3.29 & 3.44 & 7.70 & \textbf{8.07} \\
MAI-UI-2B
 & 0.48 & 3.81 & 5.03 & \textbf{7.08}
 & 4.62 & 2.57 & 4.28 & \textbf{6.96}
 & 4.78 & 2.33 & 5.56 & \textbf{7.50}
 & 3.29 & 2.90 & 4.96 & \textbf{7.18} \\
GELab-Zero-4B
 & 0.54 & 1.63 & 3.33 & \textbf{7.08}
 & 2.00 & 1.43 & \textbf{2.28} & 1.66
 & 3.36 & 3.75 & 10.21 & \textbf{13.18}
 & 1.97 & 2.27 & 5.27 & \textbf{7.31} \\
\bottomrule
\end{tabular}}
\end{table}

\paragraph{Setup.}
We assess how task-level fine-tuning changes prior-induced gains relative to the original backbones and isolate the additional contribution of explicit reliability-label supervision within CoT training. Under in-domain settings across three datasets, we compare \textbf{Ori.}, \textbf{SFT}, \textbf{Part-CoT}, and \textbf{Ours} to separate ordinary fine-tuning from explicit prior-assessment supervision. Part-CoT and Ours are matched in inputs, Action/Tool targets, training, and evaluation, differing only in explicit reliability-label supervision. We report mean $\Delta$ over all prior conditions.\looseness=-1

\paragraph{Results.}
\emph{(i) Training-induced changes in prior utilization.}
Table~\ref{tab:ablation} shows that the base models already benefit from priors, with an overall mean $\Delta$ of $3.08$\,pp. Ours raises this mean to $7.87$\,pp, an additional $4.79$\,pp, and improves over Ori.\ in $14$ of the $15$ settings. By comparison, SFT raises the mean by only $0.26$\,pp, with increases in $9$ settings and decreases in $6$. These results demonstrate that our training further strengthens prior-induced gains beyond the original backbones' existing capabilities, whereas ordinary action fine-tuning does not consistently produce such improvements.\looseness=-1
\emph{(ii) Contribution of explicit reliability supervision.}
Ours achieves a higher mean $\Delta$ than Part-CoT in $12$ of the $15$ backbone--dataset settings, including all five backbones on CUTG, and raises the overall mean from $5.59$ to $7.87$\,pp. The three exceptions occur on CMGUI for Qwen3-VL-8B and GELab-Zero-4B, and on ChiM-Nav for Qwen3-VL-8B. Under the matched training setup, these results support the additional contribution of explicit reliability-label supervision beyond the remaining CoT supervision across most evaluated settings.\looseness=-1
% \subsection{Study across Two Versions}

% \section{Discussion}

\section{Conclusion}
We presented \textsc{SAGE}, which couples annotation-free acquisition of task-aligned GUI priors with reliability-aware learning under drift-inspired perturbations. Across five backbones and three datasets, explicit prior assessment improves prior-induced gains and cross-dataset transfer; online evaluation extends this advantage to end-to-end execution. These findings support treating GUI knowledge as imperfect priors and learning when to follow, partially use, or ignore it.

\clearpage
%\subsection*{AI use statement}
%In this work, generative models are used as components of the proposed research pipeline, as described in the paper: multimodal models assist with GUI semantic analysis and state verification during exploration, generate semantic descriptions for prior construction, produce controlled natural-language perturbations, and generate structured Thought annotations for reliability-aware training. Ground-truth actions and executable tool calls are derived from recorded trajectories rather than generated by the teacher model. We also used generative AI tools for language polishing and improving the clarity and readability of the manuscript.

%All AI-assisted code and manuscript content were reviewed and verified by the authors. We take full responsibility for the final content of the paper.

\bibliographystyle{iclr2027_conference}
\bibliography{iclr2027_conference}

%%%%%%%%%%%%%%%%%%%%%%%%%%%%%%%%%%%%%%%%%%%%%%%%%%%%%%%%%%%%

\appendix
\section{Additional Exploration Details}
\label{app:explore-detail}

The explorer constructs a UI transition graph on controlled real devices without
human demonstrations or step-by-step human action annotation. At each frontier
state, candidate controls are grouped by semantic region type, local exploration
status, and control-level deduplication. Navigation and functional regions are
prioritized, homogeneous content regions receive a limited interaction budget,
and distraction regions such as advertisements or promotional cards are assigned
zero budget when identified by the multimodal annotator.

After each executed action, the explorer attempts to recover to the parent state.
If recovery lands on an unexpected screen, the observed transition is recorded as
a graph edge rather than discarded. The system then attempts graph-based replay
from the root to the target frontier and restarts the app only when replay is not
available. This recovery-aware procedure keeps the graph executable while
reducing fragmentation caused by dynamic content, recommendations, overlays, and
loading states.

Task-aligned priors are derived by enumerating reachable paths in the graph and
using a multimodal annotator to synthesize task descriptions and edge-level
natural-language operation summaries. This process produces retrievable
task--trajectory priors without human demonstrations or manual step labels.

\section{Dataset Construction and Statistics}
\label{app:dataset}

We construct learning data from three trajectory sources. \textbf{CUTG} is built from our self-explored UI transition graphs. We enumerate
reachable paths from the graphs, filter invalid or duplicate candidates with
automatic rules, and use a strong multimodal model to annotate edge-level and
trajectory-level priors. The annotation uses the pre-action screen, the
action information, the post-action screen, and the merged trajectory
visualization. \textbf{CMGUI} is a public GUI trajectory dataset. We extract a
subset and preserve its original train/test split. \textbf{ChiM-Nav} is used
directly and converted into the same trajectory format.

For each source, we construct three prompt-style variants. \textbf{Direct}
contains no external knowledge and no reasoning requirement, and is used to train
models that directly output the final action. \textbf{No-CoT} keeps the external
prior but removes the explicit reasoning trace. \textbf{Full-CoT} keeps the
external prior and supervises a structured Thought describing how the model
should use the knowledge before outputting the action. In the main experiments,
models marked as \textbf{SFT} are fine-tuned on Direct and evaluated with
No-CoT prompts, while models marked as \textbf{Ours} are fine-tuned on Full-CoT
and evaluated with Full-CoT prompts. Thus, both settings receive the same
external priors at test time, but only -Ours is trained with explicit prior
assessment.

\begin{table}[h]
\centering
\small
\setlength{\tabcolsep}{5pt}
\caption{Dataset statistics. Direct contains one sample per decision state,
while Full-CoT and No-CoT expand each Direct sample into six knowledge
conditions. Counts are reported as train/test samples; image counts denote the
number of unique screen images used by each source.}
\label{tab:dataset-statistics}
\begin{tabular}{lrrrr}
\toprule
Source & Direct & Full-CoT & No-CoT & Images \tabularnewline
\midrule
CUTG & 1,789 / 245 & 10,734 / 1,470 & 10,734 / 1,470 & 707 \tabularnewline
CMGUI & 1,996 / 292 & 11,976 / 1,752 & 11,976 / 1,752 & 2,257 \tabularnewline
ChiM-Nav & 555 / 129 & 3,330 / 774 & 3,330 / 774 & 684 \tabularnewline
\midrule
Total & 4,340 / 666 & 26,040 / 3,996 & 26,040 / 3,996 & 3,648 \tabularnewline
\bottomrule
\end{tabular}
\end{table}

For artifact release, we provide the processed and unified-format data derived
from public sources, including CMGUI and ChiM-Nav, subject to their original
licenses and terms of use. The released JSON splits include the Direct and
Full-CoT variants, together with image manifests and materialization scripts when
direct image redistribution is restricted. The No-CoT setting is an evaluation
prompt variant generated from the same prior-conditioned samples rather than a
separate released JSON split. We do not release the self-explored CUTG data,
including raw screenshots, complete trajectories, or account-related app states,
because they may contain private or license-sensitive mobile-app content.

Across all configurations, the generated train/test files contain 65,078
samples. This number counts different supervision formats derived from the same
underlying trajectories; it should not be interpreted as the number of unique GUI
states or unique task paths.

\section{Prior Perturbation Details}
\label{app:perturbation-details}

For each trajectory-level prior, we construct one clean prior and five perturbed
priors. These variants expose the model to external knowledge with different
reliability levels. We use the following prior conditions throughout the
experiments:

\begin{itemize}
    \item \textbf{Correct} denotes the clean prior that is consistent with the
    annotated trajectory.
    \item \textbf{Local replacement} keeps a correct path prefix but replaces
    later steps with a locally incorrect branch, simulating priors that are
    partially useful but later misleading.
    \item \textbf{Noise injection} inserts irrelevant or weakly related
    operations into the path, simulating retrieval noise, dynamic distractions,
    or irrelevant interface entries.
    \item \textbf{Other endpoint path} replaces the prior with a complete path to
    a different target, simulating a structurally coherent but goal-mismatched
    prior.
    \item \textbf{Semantic rewrite} keeps the operation path unchanged but
    rewrites the knowledge description, simulating expression variation or UI
    wording changes.
    \item \textbf{Step omission} removes or truncates intermediate steps,
    simulating incomplete exploration coverage or over-compressed path summaries.
\end{itemize}

These perturbations are not intended as arbitrary text augmentation. They
represent common reliability failures of self-explored GUI priors: expression
variation, missing information, local conflict, irrelevant noise, and goal
mismatch. During training, these prior variants are aligned with the same
step-level decision process, enabling the model to learn whether a prior should
be followed, partially used, or ignored.

\section{Training and Evaluation Protocol}
\label{app:training-evaluation}

All trainable models are fine-tuned using LoRA for three epochs. We use
multimodal SFT data in ShareGPT-style format, where each sample contains the
current screen image, task instruction, action history, external prior, and the
supervised assistant response. The Full-CoT response contains three parts:
Thought, Action, and tool call. Thought contains the prior assessment decision;
Action describes the next GUI operation in natural language; and tool call gives
the executable mobile action.

For -SFT models, we train on Direct data and evaluate with No-CoT prompts. For
-Ours models, we train on Full-CoT data and evaluate with Full-CoT prompts. This
evaluation protocol ensures that both models receive external priors at test
time, while only -Ours has been trained to explicitly assess prior reliability.

The main LoRA training configuration uses LoRA rank 16, LoRA alpha 32, LoRA
dropout 0.05, and applies LoRA to all supported target modules. We use bf16
training, gradient checkpoint, a cosine learning-rate schedule, warmup ratio
0.05, and learning rate 5e-5. The maximum sequence length is 8192. We use batch
size 1 with gradient accumulation 4, adjusting the number of GPUs according to
model size.

We report step accuracy and marginal accuracy gain:
\[
\Delta =
\mathrm{Acc}(\mathrm{with\ prior}) -
\mathrm{Acc}(\mathrm{no\ prior}).
\]
A positive $\Delta$ means that the given prior condition improves action
accuracy relative to the no-prior input, while a negative $\Delta$ means that the
prior misleads the model. In cross-dataset evaluation, we exclude diagonal
train-test pairs and report only off-diagonal transfer settings.

\section{Cases of Exploration}
\label{app:case}
\begin{figure}[htbp]
    \centering
    \begin{subfigure}[b]{0.45\linewidth}
        \centering
        \includegraphics[width=\linewidth]{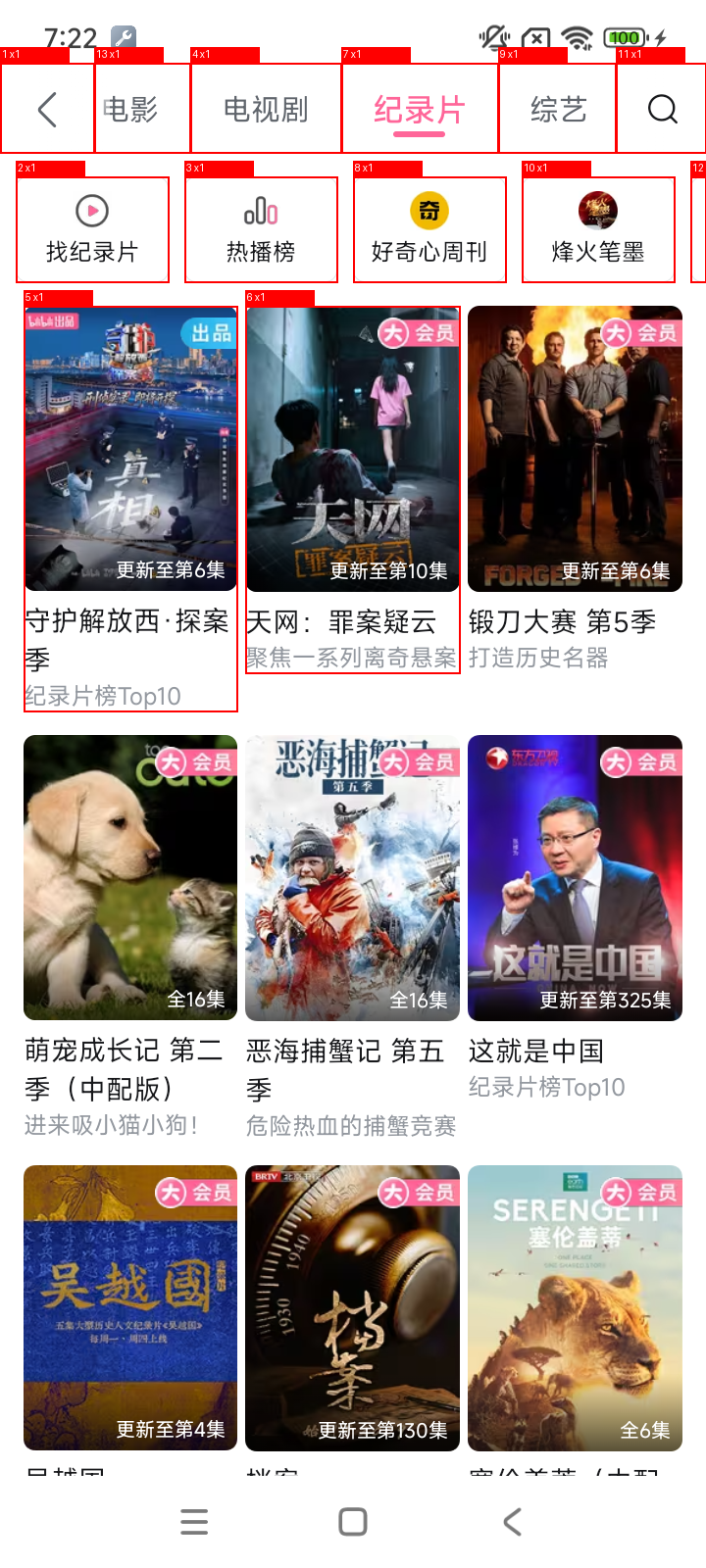}
        \caption{}
        \label{fig:left}
    \end{subfigure}
    \hfill
    \begin{subfigure}[b]{0.45\linewidth}
        \centering
        \includegraphics[width=\linewidth]{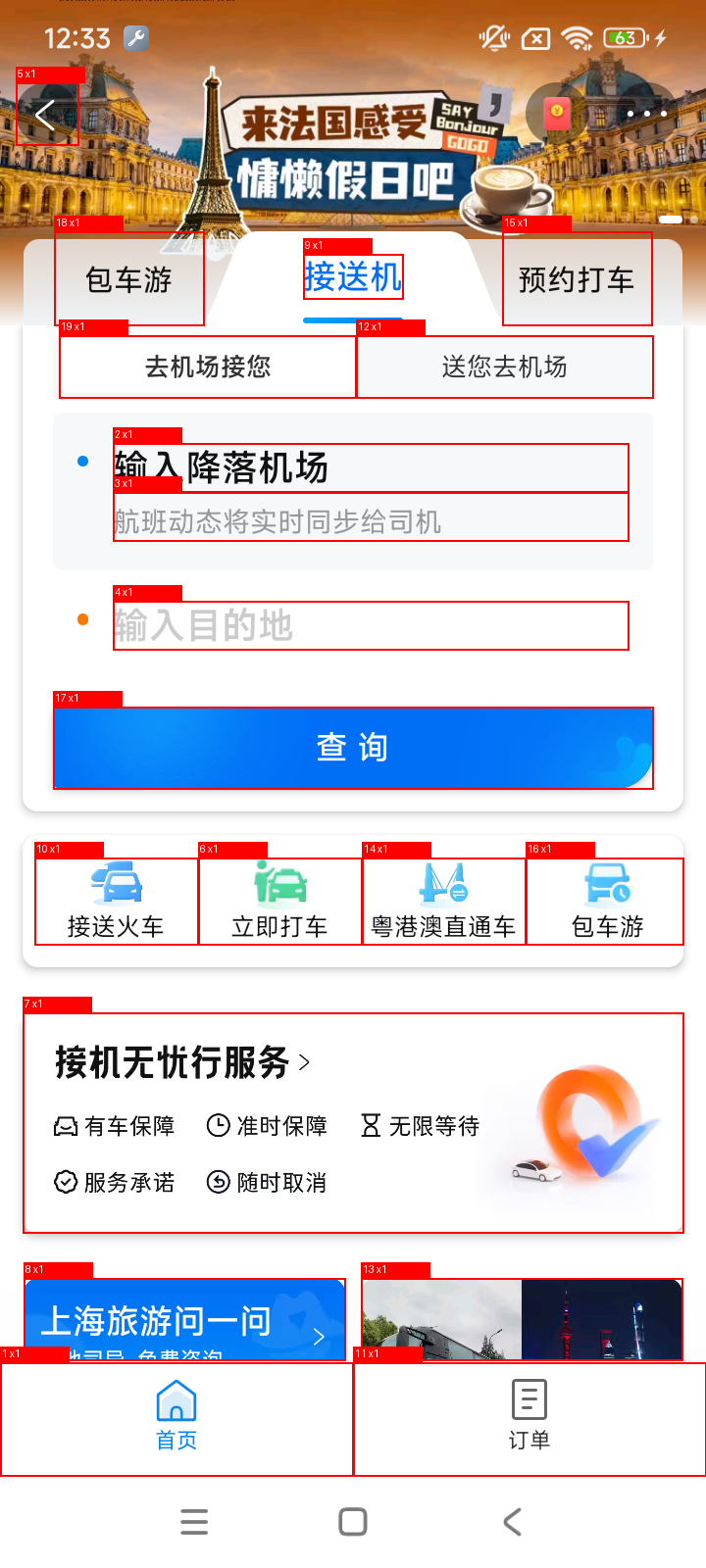}
        \caption{}
        \label{fig:right}
    \end{subfigure}
    \caption{Exploration strategies for different interface types. Red boxes indicate the controls being explored. In (a), only two homogeneous video items are explored as they lead to similar video playback interfaces; in (b), diverse functional controls are explored comprehensively.}
    \label{fig:exploration_example}
\end{figure}

As described in Sec.~\ref{sec:exploration}, during exploration, our method prioritizes interacting with controls that are likely to lead to new screens. For controls that lead to homogeneous interfaces (e.g., various product detail pages), only two are explored by design to avoid unbounded traversal of similar screens. This is illustrated in Figure~\ref{fig:exploration_example}, where red boxes indicate the controls being explored on the current screen. In subfigure (a), only two video items are explored, as they lead to similar video playback interfaces. In contrast, subfigure (b) shows a screen where different functional paths are covered more comprehensively, as the corresponding controls lead to distinct functional interfaces.

\section{Data Availability and Reproducibility}
\label{app:data-availability}

At submission time, we provide an anonymized supplementary package for review.
The package contains the core implementation, prompt templates, dataset manifests, image
materialization scripts, preprocessing scripts, and evaluation scripts needed to
inspect the pipeline and verify the benchmark format.

Upon acceptance, we will publicly release the core implementation and the
public-data portion of the benchmark. The release will include processed CMGUI
and ChiM-Nav Direct and Full-CoT JSON splits, image manifests, prompt templates,
representative training configurations, preprocessing scripts, dataset
construction scripts, and evaluation scripts. The No-CoT setting is implemented
as a prompt/evaluation variant and can be generated from the released
prior-conditioned samples rather than stored as a separate split.

To protect privacy and avoid unnecessary redistribution of sensitive operational
traces, we will not release account credentials, API keys, server addresses,
local machine paths, raw ADB dumps, raw full-device exploration logs, raw
self-explored screenshots, complete CUTG trajectories, or private app states.
For public datasets, redistribution follows the original licenses and terms of
use; when direct image redistribution is restricted, we provide image manifests
and materialization scripts instead.

\section{Statistical Significance}
\label{app:statistical-significance}

To assess the robustness of the observed gains to data splitting, we construct three
additional train--test splits under the same grouping and overlap constraints used in the
main evaluation. For each additional split, we compare Ours against SFT on four
representative backbones.

For each backbone--split setting, we estimate a $95\%$ confidence interval for the
difference in mean prior-induced accuracy gain,
$\Delta_{\mathrm{Ours}}-\Delta_{\mathrm{SFT}}$, using paired bootstrap resampling
with $B=5{,}000$. Across the three additional splits and four backbones, all 12
estimated differences are positive, and 10 of the 12 confidence intervals have
lower bounds above zero. Averaged across the three splits, the differences are
$+9.97$, $+6.60$, $+9.00$, and $+9.93$~pp for Qwen3-VL-2B, Qwen3-VL-4B,
MAI-UI-2B, and GELab-Zero-4B, respectively.

These results indicate that the average advantage of our method persists across the
tested data splits, although its magnitude varies across backbones and splits.

\section{Compute Resources}
\label{app:compute}

All experiments were conducted on a single server equipped with 8 NVIDIA RTX
4090 GPUs, each with 24GB of GPU memory. The same server was used for model
fine-tuning, inference, and evaluation.

For LoRA fine-tuning, 2B and 4B models were trained on a single RTX 4090 GPU. For
8B models, we used two RTX 4090 GPUs with DeepSpeed support. Inference and
evaluation were also performed on the same server. A 2B model requires
approximately half of one RTX 4090 GPU for serving under our evaluation
configuration, a 4B model requires one RTX 4090 GPU, and an 8B model requires two
RTX 4090 GPUs.

The reported experiments use parameter-efficient LoRA fine-tuning rather than
full-parameter training. This keeps the compute cost moderate and allows all
reported model variants to be trained and evaluated on the above 8-GPU server.
We report the hardware configuration and GPU allocation used for the experiments,
but do not report exact wall-clock runtime for each training and evaluation run.

\section{LLM Usage in the Research Pipeline}
\label{app:llm-usage}

LLMs and multimodal LLMs are used as core components of our method, not only for
writing assistance. In the exploration stage, multimodal models are used to help
analyze interface semantics and verify whether two observed screens correspond
to the same semantic interface. In prior construction, a multimodal model
generates edge-level semantic descriptions from the pre-transition screen, the
triggering action, and the post-transition screen. In prior simulation, a
language model is used to generate natural-language perturbations such as
semantic rewrites. In robust learning, a teacher multimodal model generates the
Thought field for explicit prior assessment.

The teacher model is not used to determine the final ground-truth action. The
final Action and executable tool call are reconstructed from the annotated
trajectory. This design uses the teacher model for semantic reasoning while
reducing the risk of introducing incorrect action or coordinate labels from
teacher-model hallucination.

\section{Limitations}
\label{app:limitations}

Our method has several limitations. First, the exploration module relies on
screenshots and accessibility trees. If an app provides incomplete accessibility
metadata, uses heavily customized rendering, or blocks automation, state
deduplication and control parsing may become less reliable. Second, real mobile
apps contain dynamic factors such as login states, regional content,
advertisements, A/B tests, and temporary campaigns. Although our framework treats
self-explored knowledge as imperfect priors rather than facts, severe interface
drift may still reduce retrieval quality and action grounding.

Third, our learning experiments are conducted on fixed train/test splits from
CUTG, CMGUI, and ChiM-Nav, and on a limited set of model backbones. While we
evaluate multiple model families, perturbation types, and cross-dataset transfer
settings, the results may not cover all app categories, languages, or
interaction patterns. Fourth, our training pipeline uses teacher multimodal
models to generate semantic priors and Thought annotations. This improves
scalability but may introduce teacher biases or annotation errors.

Finally, real-device exploration and teacher-model annotation introduce
non-negligible engineering and computational costs. Although our pipeline reduces
manual annotation effort, scaling it to many apps still requires stable device
execution, app-state management, model serving, and repeated quality checks.

\section{Broader Impacts and Safeguards}
\label{app:broader-impacts}

GUI agents can reduce the cost of mobile task automation, GUI testing, and GUI
data construction. Robust use of imperfect self-explored priors may improve the
reliability of agents deployed in dynamic apps, where blindly following stale or
noisy knowledge can cause failures. This may benefit accessibility tools,
software testing, and the construction of safer mobile automation systems.

The same capability also has potential risks. GUI automation could be misused
for unauthorized app interaction, spam, large-scale scraping, or actions that
violate an application's terms of service. Screenshots collected during
exploration may contain sensitive information if real personal accounts are
used. In addition, incorrect agent actions may cause unintended changes in app
state, especially when deployed outside controlled research settings.

To mitigate these risks, our experiments use controlled devices and test
accounts, avoid releasing credentials or private user data, and do not publicly
release raw self-explored screenshots. The proposed framework
does not include mechanisms for bypassing authentication, payment, access
control, or platform security restrictions. We intend the method for research on
GUI robustness, GUI testing, and controlled automation rather than unauthorized
operation of third-party applications.

\section{Asset Licenses and Released Artifacts}
\label{app:asset-licenses}

We use existing public GUI trajectory datasets, benchmarks, model backbones, and
baseline implementations only according to their stated licenses and terms of
use. For each public dataset or benchmark used in this paper, including CMGUI
and ChiM-Nav, we cite the original paper or repository and preserve the original
train/test split when applicable. The released processed data derived from these
public datasets follows the corresponding original licenses and terms of use, and
we provide attribution, source links, access dates, and preprocessing
descriptions in the released package.

We release the core implementation needed for data conversion, prior
construction, perturbation generation, prompt formatting, training, and
evaluation. The released code package includes environment specifications,
configuration files, and scripts for reproducing experiments on the public-data
portion of our benchmark. We do not release account credentials, raw private app
states, raw self-explored screenshots, or complete trajectories from our
self-explored mobile-app data. The self-explored data are used only for the
reported controlled research experiments because they may contain app content,
dynamic recommendations, advertisements, account states, or other information
requiring privacy, licensing, and terms-of-service review.

\section{App Subset for Real-Device Baseline Comparison}
\label{app: Subset}

For comparison with baseline exploration methods, we select twenty widely used mobile applications from our full explored app collection, covering online shopping, entertainment, social media, and travel scenarios. All methods are evaluated on the same app versions using three real Android devices under the same 8-hour interaction budget. Table~\ref{tab:per_app_600_steps} reports the number of unique screens and activities reached by each method at the 600th step for each app. Note that droidagent failed to reproduce on Douban.

\begin{table*}[!t]
\centering
\small
\caption{Per-app exploration results at the 600th exploration step.
\textbf{Scr.} and \textbf{Act.} denote the number of deduplicated screens and
distinct activities reached, respectively. The best value in each (app, metric)
cell is \textbf{bolded}; ``--'' marks runs that crashed before completion.}
\label{tab:per_app_600_steps}
\setlength{\tabcolsep}{4.5pt}
\renewcommand{\arraystretch}{1.2}
\footnotesize
\begin{tabular}{@{}l cc cc cc cc >{\columncolor{gray!12}}c >{\columncolor{gray!12}}c@{}}
\toprule
\multirow{2}{*}{\textbf{App}}
 & \multicolumn{2}{c}{\textbf{DroidBot}}
 & \multicolumn{2}{c}{\textbf{DroidAgent}}
 & \multicolumn{2}{c}{\textbf{LLM-Explorer}}
 & \multicolumn{2}{c}{\textbf{Humanoid}}
 & \multicolumn{2}{c}{\textbf{Ours}} \\
\cmidrule(lr){2-3}\cmidrule(lr){4-5}\cmidrule(lr){6-7}\cmidrule(lr){8-9}\cmidrule(lr){10-11}
 & Scr. & Act. & Scr. & Act. & Scr. & Act. & Scr. & Act. & Scr. & Act. \\
\midrule
Ctrip
 & 32 & 12 & 11 & 8 & 60 & 1 & 16 & 4
 & \textbf{219} & \textbf{19} \\
Bilibili
 & 34 & 9 & 30 & 16 & 56 & 10 & 54 & 15
 & \textbf{196} & \textbf{45} \\
NetEase Cloud Music
 & 19 & 12 & 2 & 3 & 84 & 1 & 40 & 22
 & \textbf{188} & \textbf{34} \\
Amap
 & 10 & 2 & 11 & 2 & 103 & \textbf{6} & 17 & 2
 & \textbf{142} & 2 \\
Douban
 & 44 & 30 & -- & -- & 45 & 8 & 69 & 38
 & \textbf{120} & \textbf{52} \\
Kuaishou
 & 62 & 15 & 26 & 6 & 10 & 4 & 45 & 15
 & \textbf{208} & \textbf{37} \\
Taobao
 & 19 & 9 & 12 & 8 & 61 & 8 & 44 & \textbf{18}
 & \textbf{103} & 15 \\
Xiaohongshu
 & 39 & 16 & 37 & 25 & 66 & 8 & 52 & 18
 & \textbf{176} & \textbf{29} \\
Ximalaya
 & 11 & 2 & 30 & 2 & \textbf{88} & \textbf{7} & 42 & 6
 & 63 & 3 \\
Qunar Travel
 & 17 & 9 & 5 & 7 & 65 & 9 & 31 & 11
 & \textbf{154} & \textbf{14} \\
iQIYI
 & 25 & 10 & 37 & 12 & 25 & 10 & 66 & 11
 & \textbf{147} & \textbf{24} \\
Tencent Video
 & 33 & 11 & 0 & 1 & 38 & 10 & 32 & 12
 & \textbf{140} & \textbf{15} \\
Youdao Dictionary
 & 38 & 15 & 14 & 5 & 52 & 11 & 73 & 26
 & \textbf{138} & \textbf{47} \\
DiDi
 & 31 & 8 & 11 & 4 & 11 & 5 & 25 & 7
 & \textbf{117} & \textbf{19} \\
YOUKU
 & 5 & 4 & 55 & \textbf{21} & 65 & 9 & 61 & 13
 & \textbf{117} & 16 \\
QQ
 & 55 & 13 & 38 & 11 & 58 & 6 & 60 & 16
 & \textbf{192} & \textbf{36} \\
Baidu Netdisk
 & 20 & 7 & 19 & 8 & 51 & 9 & 45 & 4
 & \textbf{145} & \textbf{49} \\
Quark
 & 15 & 2 & 5 & 2 & 47 & \textbf{6} & 20 & 3
 & \textbf{150} & 2 \\
DingTalk
 & 38 & 33 & 11 & 6 & 92 & 14 & 46 & 18
 & \textbf{134} & \textbf{80} \\
Fanqie Changting
 & 9 & 4 & 8 & 3 & 9 & 4 & 33 & 7
 & \textbf{113} & \textbf{34} \\
\midrule
\textbf{Average}
 & 27.8 & 11.2 & 19.1 & 7.9 & 54.3 & 7.3 & 43.6 & 13.3
 & \textbf{148.1} & \textbf{28.6} \\
\bottomrule
\end{tabular}
\end{table*}

\section{Loss-Weight Sensitivity}
\label{app:loss-weight-sensitivity}
\begin{table}[!htbp]
    \centering
    \small
    \caption{Loss-weight sensitivity on CUTG. Values are mean accuracy gains (pp) over each model's own no-prior baseline, averaged over six prior conditions. Bold indicates the best result per backbone.}
    \label{tab:loss-weight-sensitivity}
    \begin{tabular}{lrr}
        \toprule
        Thought:Action:Tool & Qwen3-VL-2B & Qwen3-VL-4B \\
        \midrule
        $1{:}1{:}1$ (default)
            & \textbf{+11.70} & \textbf{+11.23} \\
        $2{:}1{:}1$ & +9.52  & +8.91  \\
        $1{:}2{:}1$ & +10.51 & +9.13  \\
        $1{:}1{:}2$ & +8.76  & +10.05 \\
        \bottomrule
    \end{tabular}
\end{table}

\paragraph{Setup.}
We examine the sensitivity of Full-CoT training to the $Thought:Action:Tool$ loss weights on CUTG using Qwen3-VL-2B and Qwen3-VL-4B. We compare the default ratio $1{:}1{:}1$ with three alternatives that double the weight of one response component, while keeping the data, train--test split, and all other training settings fixed. For each setting, we report the mean prior-induced accuracy gain,
$\overline{\Delta} = \frac{1}{6}\sum_{k=1}^{6} \left[ \mathrm{Acc}_{k}-\mathrm{Acc}_{\mathrm{no\text{-}prior}} \right]$, where the no prior accuracy is measured separately for each trained model.

\paragraph{Results.} Table~\ref{tab:loss-weight-sensitivity} shows positive mean gains under all tested weight settings for both backbones. The default $1{:}1{:}1$ ratio yields the highest mean gain in each case. Doubling an individual component weight reduces the gain by $1.19$ to $2.94$ pp for Qwen3-VL-2B and $1.17$ to $2.31$ pp for Qwen3-VL-4B. Thus, positive prior-induced gains persist under these local weight changes, although their magnitude varies. These results support the default ratio within the tested configurations; they do not establish a globally optimal weighting or a ranking by absolute action accuracy.

\section{Reliability Assessment on Held-Out Real Drift}
\label{app:real-drift-reliability}
\begin{table}[t]
\centering
\small
\caption{
Reliability assessment on held-out real cross-version GUI drift.
We report class-wise F1 and Macro-F1 (\%) against manually annotated
\textsc{Follow}/\textsc{Partial}/\textsc{Ignore} labels.
Best results in each column are bolded.
}
\label{tab:real-drift-reliability}
\setlength{\tabcolsep}{7pt}
\renewcommand{\arraystretch}{1.08}
\begin{tabular}{lcccc}
\toprule
Backbone &
\textsc{Follow} F1 &
\textsc{Partial} F1 &
\textsc{Ignore} F1 &
Macro-F1 \\
\midrule
GELab-Zero-4B
& \textbf{92.09}
& \textbf{64.08}
& 64.97
& \textbf{73.71} \\

MAI-UI-2B
& 91.40
& 56.57
& 60.24
& 69.41 \\

Qwen3-VL-2B
& 91.33
& 54.55
& \textbf{65.54}
& 70.47 \\

Qwen3-VL-4B
& 90.00
& 56.41
& 58.62
& 68.34 \\
\midrule
Average
& 91.21
& 58.82
& 63.57
& 70.48 \\
\bottomrule
\end{tabular}
\end{table}

\paragraph{Setup.}
To directly evaluate reliability assessment under real GUI drift, we construct
a held-out cross-version drift set that is not used for training.
Each prior is manually annotated with one of the three reliability labels
\textsc{Follow}, \textsc{Partial}, or \textsc{Ignore}.
We then compare the reliability decision generated by each trained model
against these human annotations and report class-wise F1 and Macro-F1.
Unlike the controlled perturbation evaluation in Sec.~4.2, these labels are
derived from real cross-version drift rather than from the synthetic
perturbation operator that generated the prior.

\paragraph{Results.}
Table~\ref{tab:real-drift-reliability} shows that all four backbones achieve
at least $90\%$ F1 on \textsc{Follow}, with an average of $91.21\%$.
The average F1 scores for \textsc{Partial} and \textsc{Ignore} are
$58.82\%$ and $63.57\%$, respectively, yielding an average Macro-F1 of
$70.48\%$ across backbones.
\textsc{Partial} is consistently the most challenging category, whereas
\textsc{Follow} is identified reliably across all models.
These results provide direct evidence that the learned reliability decisions
transfer to held-out real cross-version drift rather than only to the
synthetic perturbation taxonomy used during training.

% \section{Technical appendices and supplementary material}
% Technical appendices with additional results, figures, graphs, and proofs may be submitted with the paper submission before the full submission deadline (see above). You can upload a ZIP file for videos or code, but do not upload a separate PDF file for the appendix. There is no page limit for the technical appendices. 

% Note: Think of the appendix as ``optional reading'' for reviewers. The paper must be able to stand alone without the appendix; for example, adding critical experiments that support the main claims to an appendix is inappropriate. 

% \begin{table}[t]
% \centering
% \small
% \setlength{\tabcolsep}{5pt}
% \caption{Dataset statistics. Direct contains one sample per decision state,
% while Full-CoT and No-CoT expand each Direct sample into six knowledge
% conditions. Counts are reported as train/test samples; image counts denote the
% number of unique screen images used by each source.}
% \label{tab:dataset_statistics}
% \begin{tabular}{lrrrr}
% \toprule
% Source & Direct & Full-CoT & No-CoT & Images \tabularnewline
% \midrule
% Ours & 1,789 / 245 & 10,734 / 1,470 & 10,734 / 1,470 & 707 \tabularnewline
% CMGUI & 1,996 / 292 & 11,976 / 1,752 & 11,976 / 1,752 & 2,257 \tabularnewline
% ChiM-Nav & 555 / 129 & 3,330 / 774 & 3,330 / 774 & 684 \tabularnewline
% \midrule
% Total & 4,340 / 666 & 26,040 / 3,996 & 26,040 / 3,996 & 3,648 \tabularnewline
% \bottomrule
% \end{tabular}
% \end{table}

%%%%%%%%%%%%%%%%%%%%%%%%%%%%%%%%%%%%%%%%%%%%%%%%%%%%%%%%%%%%

\end{document}